\documentclass[10pt,twocolumn,letterpaper]{article}

\usepackage[pagenumbers]{wacv} 

\definecolor{wacvblue}{rgb}{0.21,0.49,0.74}
\usepackage[pagebackref,breaklinks,colorlinks,allcolors=wacvblue]{hyperref}

\title{MorphoSHAP: Rethinking the Unit of Attribution in Explanation for Deep Visual Models}

\author{
\textbf{Anirudh Prabhakaran}, \textbf{Alexandre Rocchi}, \textbf{Gianni Franchi}\\
AMIAD, Pôle Recherche, Palaiseau\\
{\tt\small Corresponding author: anirudh.prabhakaran@ip-paris.fr}
}

\usepackage{xspace}
\usepackage{bbm}
\usepackage{multirow}
\usepackage{graphicx}
\usepackage{xcolor}
\usepackage{tikz}
\usetikzlibrary{positioning,calc,arrows.meta}

\newcommand{\method}{\textsc{MorphoSHAP}\xspace}

\begin{document}
\maketitle

\begin{abstract}
Visual attribution methods typically explain predictions using pixels,
superpixels, or regular patches. These representations can localize important
regions, but provide limited information about their structure.
We introduce \method, a model-agnostic post-hoc method that instead uses
\emph{morphological shapes} as the players of a Shapley attribution game.
Using the Tree of Shapes, each shape is described by its scale, geometry, and
signed contribution, providing explanations of \emph{where} the evidence lies,
\emph{what type of structure} carries it, and \emph{how strongly} it affects
the prediction.
This shared morphological vocabulary enables spatial, textual, and global
class-level explanations beyond image-specific heatmaps.
To the best of our knowledge, \method is the first SHAP-based image attribution
framework to combine these different forms of explanation.
Across five diverse datasets and three architectures, \method achieves strong
insertion/deletion performance and outperforms competing attribution methods
on several benchmarks.
Finally, a user study shows that \method provides explanations that are easy
to use and are preferred over standard attribution baselines.

\end{abstract}

\footnote{anirudh.prabhakaran@ip-paris.fr}

\section{Introduction}
\label{sec:intro}

\begin{figure}[t]
    \centering
    \includegraphics[width=0.73\columnwidth]{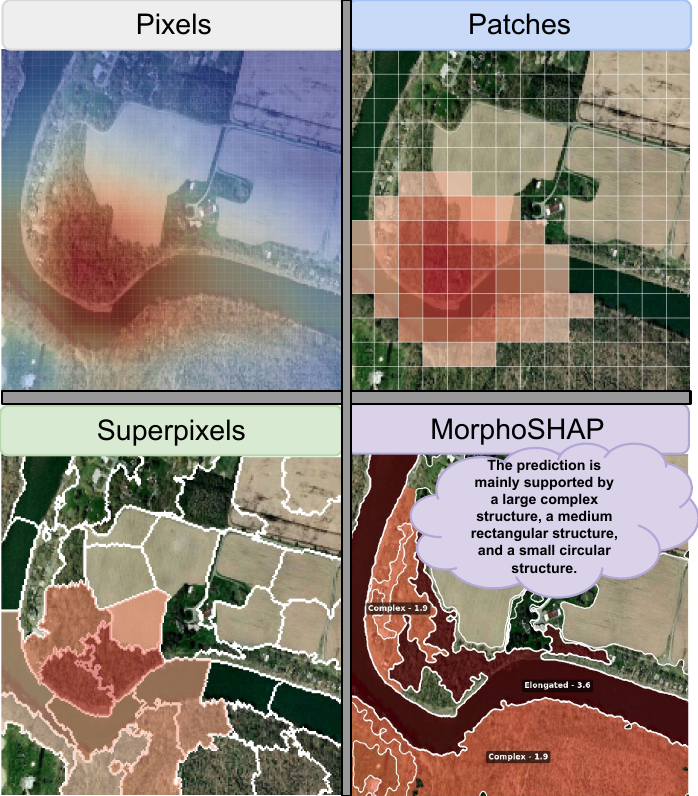}
\caption{
\textbf{Rethinking the unit of visual attribution.}
While conventional methods explain predictions through pixels, patches, or
superpixels, \method uses morphological shapes as Shapley players.
}
    \label{fig:morphoshap_comparison}
\end{figure}

Understanding why a Deep Neural Network (DNN) makes a prediction remains a
central challenge in explainable artificial intelligence (XAI).
In computer vision, attribution methods assign importance scores to parts of
an image to identify the evidence supporting or opposing a prediction.
Existing approaches include pixel-level saliency
maps~\cite{simonyan2013deep,sundararajan2017axiomatic,bilgicc2026disentangling,bousselham2025legrad,chefer2021transformer,chefer2021generic,kazmierczak2025enhancing},
activation-based explanations~\cite{selvaraju2017grad}, local surrogate
methods~\cite{ribeiro2016should}, and Shapley-based
attributions~\cite{lundberg2017unified}.
Despite their differences, these methods share the same goal:
determining \emph{which elements of the image matter for the prediction}.

This raises a more basic question:
\emph{what should be the elementary unit of visual attribution?}
Pixel-based methods use individual pixels, perturbation methods often rely on
superpixels, and vision transformers naturally suggest regular patches.
However, this choice directly defines the space in which the explanation is
expressed.
Changing the elementary units changes the players of the attribution problem,
and therefore changes the explanation itself.
\textit{We argue that the choice of the explanatory representation should be treated
as a central design decision in visual XAI.}

Pixels, superpixels, and patches are useful representations, but they do not
necessarily correspond to interpretable visual structures.
Pixels are highly localized but have little meaning in isolation.
Superpixels provide spatially coherent regions, but may split one structure or
merge several structures.
Regular patches are convenient computational units, but their boundaries are
defined by a grid rather than by image geometry.
As a result, conventional attribution maps can tell us \emph{where} important
evidence is located, but provide limited information about \emph{what type of
visual structure} the model relies on.

In this work, we investigate a different representation:
\emph{morphological shapes as elementary units of attribution}.
Images contain connected structures and shapes at different scales that can be
described through simple geometric properties such as size, elongation,
circularity, or rectangularity.
These properties are both spatially meaningful and nameable.
Shape therefore provides a natural bridge between the localization offered by
attribution maps and a structured description of the visual evidence used by
the model.

Mathematical morphology~\cite{serra1983image} provides a natural framework for
extracting such structures.
Based on this idea, we introduce \method, a model-agnostic post-hoc framework
for \emph{shape-based attribution}.
Given an image, \method uses the Tree of
Shapes~\cite{ballester2003tree,carlinet2018tree,geraud2022proof} to extract a
hierarchical set of morphological shapes.
Each shape is then characterized by its scale and geometric type, yielding
descriptions such as \emph{small circle}, \emph{medium rectangle}, or
\emph{large elongated structure}.
These morphological shapes, rather than pixels or superpixels, are finally
used as the players of a cooperative game, and their contribution to the
prediction is estimated using Shapley values~\cite{lundberg2017unified}.
The main difference with standard Shapley-based image explanations therefore
lies not in the attribution rule itself, but in the
\emph{representation on which the Shapley game is defined}.

This representation provides richer explanations than a conventional
attribution map.
For each image, \method associates every relevant structure with its spatial
support, scale, geometric type, and signed contribution.
The same explanation can therefore be shown as a heatmap or with a classical SHAP plot (Waterfall, or Beeswarm plot, ...) or expressed in
words, e.g.,
\emph{``the prediction is mainly supported by a large elongated structure and
a small circular region.''}
Moreover, because scale and geometric labels have the same meaning across
images, these explanations can be aggregated over a dataset.
This allows us to study, for example, whether a class is systematically
associated with small circular structures, medium rectangles, or large
elongated shapes.
Thus, the same morphological vocabulary supports both local and global model
analysis.

Our contributions are summarized as follows:
\begin{itemize}

    \item
    We introduce \method, a model-agnostic post-hoc method that uses
    \emph{morphological shapes} as the elementary units of a Shapley attribution
    game.

    \item
    We define a structured morphological explanation in which each shape is
    described by its \emph{scale}, \emph{geometry}, and \emph{signed contribution}.

    \item This shared vocabulary enables \emph{local}, \emph{textual}, and
\emph{global} explanations beyond standard heatmaps.
To the best of our knowledge, \method is the first SHAP-based image attribution
framework to provide this combination of spatial, geometric, textual, and
global explanations.
 
    \item
    Experiments across datasets and architectures show that \method provides
    faithful shape-level attributions with more structured and interpretable
    explanations than pixel-, patch-, and superpixel-based alternatives.

\end{itemize}

\section{Related Work}
\label{sec:related-work}

\subsection{Gradient-Based Attribution}

Gradient-based methods such as Grad-CAM~\cite{selvaraju2017grad}, Grad-CAM++~\cite{chattopadhay2018grad}, Layer-CAM~\cite{jiang2021layercam}, Score-CAM~\cite{wang2020score} and ShapleyCAM~\cite{cai2025cams} exploit model gradients or activation maps to produce spatial heatmaps. These techniques are computationally efficient and widely adopted for visualizing classifier decisions. However, they are inherently model-dependent: they require access to the internal layer activations and gradients, and they lack precise object boundaries, often highlighting diffuse regions rather than distinct entities. Furthermore, gradient saturation and reliance on specific architectural choices limit their applicability across diverse model families.

\subsection{Hierarchical and Structured SHAP}

Several works adapt Shapley-based attribution to structured image
representations in order to reduce the cost and improve the coherence of visual
explanations~\cite{lundberg2017unified,teneggi2022fast,rashid2026shapbpt,
hasic2026aa,hasic2025superpixel,he2026sharpen}.
Hierarchical methods such as h-Shap~\cite{teneggi2022fast} and
ShapBPT~\cite{rashid2026shapbpt} organize image regions into trees and compute
hierarchical Shapley/Owen-style contributions, while superpixel-based methods
exploit region correlations or affinities to reduce the attribution
game~\cite{hasic2025superpixel,hasic2026aa}.
Other approaches use learned segmentation or concepts, e.g.,
Explain Any Concept~\cite{sun2023explain}, which relies on
SAM~\cite{kirillov2023segment} to define interpretable image regions.
In contrast, \method defines the Shapley players directly from a
multi-scale morphological hierarchy, providing each region with an explicit
scale and geometric description without relying on a regular partition or an
external segmentation model.

\subsection{Mathematical Morphology for XAI}
Mathematical morphology provides a principled framework for describing image
structures through shape, connectivity, and scale~\cite{serra1983image,bouchet2016fuzzy},
and has been widely used to characterize visual patterns in microscopic and
remote-sensing imagery~\cite{franchi2018enhanced,franchi2016morphological,
franchi2014comparative,cavallaro2016spectral,cavallaro2017automatic}.
Morphological operators have also been integrated into neural
networks~\cite{masci2013learning,saeedan2018detail,mondal2019dense,FRANCHI2020107246}, mainly as feature-extraction or architectural
components rather than as post-hoc explanatory units. There have been multiple morphological decompositions; in particular, the Tree of Shapes provides a self-dual hierarchical
decomposition of an image into nested connected structures at multiple
scales~\cite{ballester2003tree,carlinet2018tree,geraud2022proof}.
In contrast to prior uses of morphology, \method exploits these structures
directly as the units of attribution, linking mathematical morphology with XAI.


\section{Method}
\label{sec:method}

\begin{figure*}[t]
	\centering
	\includegraphics[width=\linewidth]{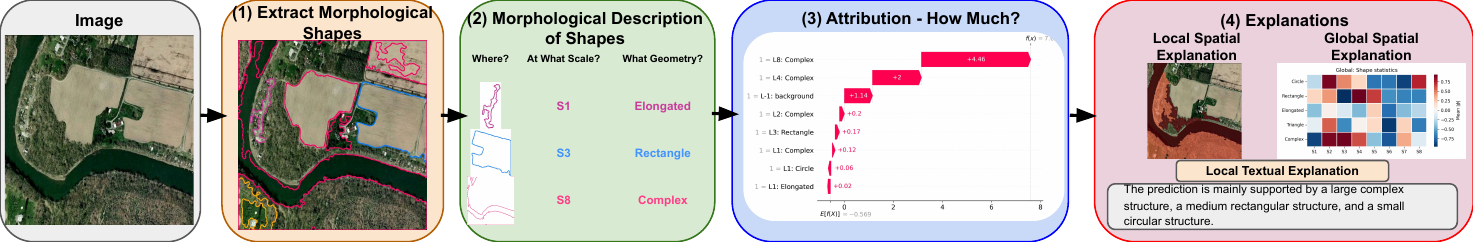}
	\caption{Overview of the MorphoSHAP explanation pipeline. \textbf{1) Extraction:} The input image is decomposed into hierarchical morphological regions using the Tree of Shapes. \textbf{2) Semantic Description:} Each isolated region is mathematically evaluated and assigned a semantic tag (e.g., Triangular, Elongated, Complex). \textbf{3) Attribution:} A region-based Shapley value estimator quantifies the precise contribution of each shape to the model's prediction. \textbf{4) Explanation:} The framework outputs rich, multi-modal explanations, including local spatial heatmaps, structured textual summaries, and global class-wise shape distributions.}
	\label{fig:pipeline_overview}
\end{figure*}


\subsection{Overview and Problem Formulation}
\label{sec:method_overview}

Let \(x\in\mathbb{R}^{H\times W\times C}\) be an input image and
\(f:\mathbb{R}^{H\times W\times C}\rightarrow\mathbb{R}^{K}\) a black-box
classifier. For a target class \(y\), we explain the corresponding model
output \(f_y(x)\), taken as the class logit unless stated otherwise.

The key idea of \method is to define attribution over \emph{morphological
shapes}, rather than pixels, superpixels, or regular patches.
Given \(x\), we extract \(M\) morphological shapes and progressively enrich
each shape \(b_i\) with its scale \(s_i\), geometric category \(g_i\), and
estimated contribution \(\hat{\phi}_i\):
\begin{equation}
    b_i
    \longrightarrow
    (b_i,s_i)
    \longrightarrow
    (b_i,s_i,g_i)
    \longrightarrow
    (b_i,s_i,g_i,\hat{\phi}_i).
    \label{eq:morphoshap_pipeline}
\end{equation}
The resulting explanation is
\begin{equation}
    \mathcal{E}_y(x)
    =
    \left\{
        (b_i,s_i,g_i,\hat{\phi}_i)
    \right\}_{i=1}^{M}.
    \label{eq:full_explanation}
\end{equation}
Each explanatory unit therefore specifies \emph{where} the evidence lies,
\emph{at what scale}, \emph{with which geometry}, and \emph{how strongly} it
contributes to the prediction.


\subsection{Morphological Shape Decomposition}
\label{sec:tree_of_shapes}

\paragraph{Tree of Shapes.}
Among morphological decompositions~\cite{ballester2003tree}, we use the Tree of Shapes because it
directly represents an image as a hierarchy of connected shapes.
Although extensions to color images exist~\cite{carlinet2015mtos}, we present
the formulation for a grayscale image $u$ for simplicity.

\paragraph{Upper and lower level sets.}For a gray level $\lambda$, the lower and upper level sets are
\begin{align}
    L_{\lambda}(u)
    &=
    \{p\in\Omega : u(p)<\lambda\},
    \\
    U_{\lambda}(u)
    &=
    \{p\in\Omega : u(p)\geq\lambda\}.
\end{align}
Their connected components capture dark and bright structures at different
intensity levels~\cite{ballester2003tree,carlinet2018tree,geraud2022proof}.
The Tree of Shapes combines both families into a self-dual representation.

\paragraph{Shapes and hierarchy.}
To obtain shapes, holes inside each connected component $\Gamma$ are filled
through the saturation operator
\begin{equation}
    \operatorname{Sat}(\Gamma)
    =
    \Omega
    \setminus
    \operatorname{CC}
    \left(
        \Omega\setminus\Gamma,
        p_{\infty}
    \right),
    \label{eq:saturation}
\end{equation}
where $\operatorname{CC}(A,p)$ denotes the connected component of $A$
containing $p$.
The resulting family of shapes is
\begin{equation}
    \mathcal{S}(u)
    =
    \mathcal{S}^{<}(u)
    \cup
    \mathcal{S}^{\geq}(u),
    \label{eq:tos_shapes}
\end{equation}
where $\mathcal{S}^{<}(u)$ and $\mathcal{S}^{\geq}(u)$ are obtained from the
lower and upper level sets, respectively.
Two shapes are either disjoint or nested, and the inclusion relation
\begin{equation}
    b_i \preceq b_j
    \quad\Longleftrightarrow\quad
    b_i \subseteq b_j
\end{equation}
organizes them into the Tree of Shapes.

\paragraph{Morphological explanatory units.}
We use these structures as the explanatory units of \method:
\begin{equation}
    \mathcal{B}(x)
    =
    \{b_1,\ldots,b_M\},
    \qquad
    b_i\in\mathcal{S}(u),
    \label{eq:blob_registry}
\end{equation}
excluding the root corresponding to the full image.
Each shape is represented by a binary mask
\begin{equation}
    m_i(p)
    =
    \mathbf{1}_{\{p\in b_i\}},
    \qquad
    m_i\in\{0,1\}^{H\times W}.
\end{equation}
Unlike a flat segmentation or regular patch grid, this representation preserves
the hierarchy of image structures, allowing fine shapes to be nested inside
larger ones.


\subsection{Multi-Scale Morphological Characterization}
\label{sec:scale_characterization}

The Tree of Shapes provides the morphological structures, which we further
characterize by scale. For each shape $b_i$, we compute its relative area
\(
    \rho_i = \frac{|b_i|}{HW},
\)
and assign it to one of eight ordered scale levels
$\mathcal{S}_{\mathrm{scale}}=\{S_1,\ldots,S_8\}$, from fine to coarse:
\[
s_i=S_k
\quad\Longleftrightarrow\quad
\tau_{k-1}\leq\rho_i<\tau_k,
\]
where $0=\tau_0<\cdots<\tau_8=1$.
The exact scale intervals are reported in Appendix~\ref{Appendix:Extra_info_scale}.
The resulting representation is
\begin{equation}
    \mathcal{B}_{\mathrm{scale}}(x)
    =
    \{(b_i,s_i)\}_{i=1}^{M}.
    \label{eq:shape_scale}
\end{equation}


\subsection{Geometric Characterization and Shape Naming}
\label{sec:shape_characterization}

\paragraph{Geometric Shape Vocabulary.} Scale alone cannot describe the geometry of a structure.
We therefore associate each morphological shape with a discrete geometric
label
\(
    g_i
    \in
    \mathcal{G},
\)
where
\begin{equation}
    \begin{aligned}
        \mathcal{G} = \{&
        \texttt{Elongated},\,
        \texttt{Circle},\,
        \texttt{Triangle},\\
        &\texttt{Rectangle},\,
        \texttt{Polygon},\,
        \texttt{Complex}
        \}.
    \end{aligned}
    \label{eq:shape_vocabulary}
\end{equation}
We investigate two mechanisms for assigning these names:
(i) a deterministic rule-based geometric classifier, and
(ii) a learned classifier.
Both operate on the same geometric information and produce labels from the
same vocabulary $\mathcal{G}$.


\paragraph{Geometric descriptors.}

For every shape $b_i$, we extract its external contour and compute a compact
descriptor vector
\begin{equation}
    d_i
    =
    \left[
        A_i,\,
        P_i,\,
        \operatorname{AR}_i,\,
        C_i,\,
        Q_i,\,
        n_i
    \right],
    \label{eq:geometric_descriptor}
\end{equation}
where $A_i$ is the area, $P_i$ the perimeter,
$\operatorname{AR}_i$ the aspect ratio of its minimum-area bounding rectangle,
$C_i$ its circularity,
$Q_i$ its solidity, and $n_i$ the number of vertices obtained from a polygonal
approximation.


\paragraph{Rule-based geometric naming.}

Our first strategy is entirely deterministic.
It applies a sequence of geometric rules to $d_i$.
The order of the rules is intentional and makes the resulting categories
mutually exclusive. See Appendix \ref{Appendix:Extra_info_naming} for more information on the Rule-based geometric naming.

We denote this deterministic mapping by
\begin{equation}
    g_i
    =
    h_{\mathrm{rule}}(d_i).
    \label{eq:rule_classifier}
\end{equation}


\paragraph{Learned geometric naming.}

The hand-designed thresholds above provide a transparent and fully
deterministic naming mechanism.
However, fixed geometric rules may not always capture the interaction between
different descriptors.
We therefore also consider a data-driven variant. Let
\(
    h_{\mathrm{DT}}(\cdot;\psi):
    \mathbb{R}^{D}
    \rightarrow
    \mathcal{G}
\)
denotes an MLP classifier parameterized by $\psi$.
It receives an enhanced geometric descriptor vector $d_i$ and predicts
\begin{equation}
    g_i
    =
    h_{\mathrm{DT}}(d_i;\psi).
    \label{eq:decision_tree}
\end{equation}

The training data, hyperparameters, and classifier information used are
reported in the Appendix \ref{Appendix:Extra_info_Learned}. Appendix  \ref{Appendix:Extra_info_comparisons} compares the two naming strategies. Both mechanisms lead to the same structured representation
\begin{equation}
    \mathcal{B}_{\mathrm{morph}}(x)
    =
    \left\{
        (b_i,s_i,g_i)
    \right\}_{i=1}^{M}.
    \label{eq:morphological_representation}
\end{equation}


\subsection{Morphological Shape Contribution}
\label{sec:shape_shapley}

The previous stages define \emph{what} the explanatory units are.
We now quantify \emph{how much} each morphological shape contributes to the
prediction.
Each shape $b_i\in\mathcal{B}(x)$ is treated as a player in a cooperative
game.
Its scale $s_i$ and geometric label $g_i$ only describe the player; the model
is perturbed through the spatial mask $m_i$.

\paragraph{Coalitions of morphological shapes.}
Let $S\subseteq\mathcal{B}(x)$ be a coalition of active shapes, with support
\begin{equation}
    m_S
    =
    \bigvee_{b_i\in S}m_i,
    \label{eq:coalition_mask}
\end{equation}
where $\vee$ denotes the pixel-wise logical OR.
Because the Tree of Shapes is hierarchical, shapes may be nested, and $m_S$
keeps every pixel belonging to at least one active shape.

Let
\begin{equation}
    m_{\mathcal{B}}
    =
    \bigvee_{i=1}^{M}m_i,
    \qquad
    m_{\mathrm{bg}}
    =
    1-m_{\mathcal{B}}.
\end{equation}
Using a constant mid-gray reference image $x_0$  to represent an absent shape,
the image associated with coalition $S$ is
\begin{equation}
\begin{split}
    x_S
    =\;&
    m_{\mathrm{bg}}\odot x
    +
    m_S\odot x \\
    &+
    \left(m_{\mathcal{B}}-m_S\right)\odot x_0,
\end{split}
    \label{eq:coalition_image}
\end{equation}
where $\odot$ denotes element-wise multiplication.
Active shapes therefore keep their original pixels, inactive shapes are
replaced by $x_0$, and the residual background remains unchanged.
The coalition value for class $y$ is
\begin{equation}
    v_y(S)=f_y(x_S).
    \label{eq:value_function}
\end{equation}

\paragraph{Shapley contribution.}
The Shapley value of shape $b_i$ is its average marginal contribution over all
coalitions of the remaining shapes:
\begin{equation}
    \phi_i
    =
    \sum_{
        S\subseteq
        \mathcal{B}\setminus\{b_i\}
    }
    \frac{|S|!(M-|S|-1)!}{M!}
    \left[
        v_y(S\cup\{b_i\})
        -
        v_y(S)
    \right].
    \label{eq:exact_shapley}
\end{equation}
Positive values support class $y$, while negative values oppose it.
Since exact computation requires an exponential number of coalitions, we use
KernelSHAP~\cite{lundberg2017unified}, which samples coalitions and fits a
Shapley-weighted additive surrogate whose coefficients
$\hat{\phi}_i$ estimate the contribution of each morphological shape.


\subsection{Local Morphological Explanations}
\label{sec:local_explanation}

For a single image, \method produces a structured explanation
\begin{equation}
    \mathcal{E}_y(x)
    =
    \left[
        (b_i,s_i,g_i,\hat{\phi}_i)
    \right]_{i=1}^{M},
\end{equation}
which we rank according to $|\hat{\phi}_i|$.

This representation supports three complementary visualizations.

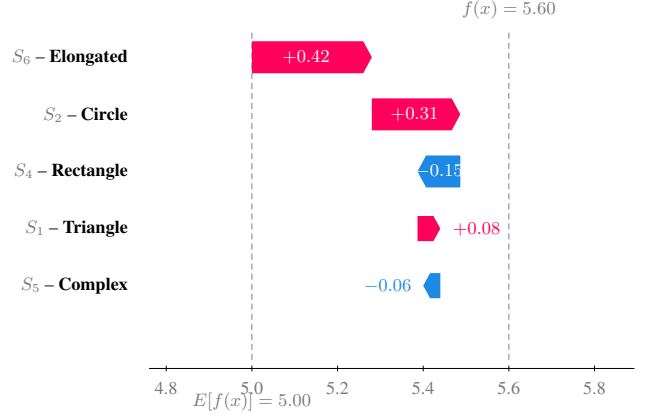
\begin{figure}[t]
    \centering

    \resizebox{\linewidth}{!}{%
    \begin{tikzpicture}[font=\small]

        \definecolor{poscol}{RGB}{255,0,90}
        \definecolor{negcol}{RGB}{30,136,229}
        \definecolor{graytxt}{RGB}{120,120,120}

        \def\yA{6.0}
        \def\yB{5.0}
        \def\yC{4.0}
        \def\yD{3.0}
        \def\yE{2.0}

        \draw[black, line width=0.5pt]
            (2.7,0.55) -- (11.2,0.55);

        \foreach \x/\lab in {
            3.0/4.8,
            4.5/5.0,
            6.0/5.2,
            7.5/5.4,
            9.0/5.6,
            10.5/5.8
        }{
            \draw[black, line width=0.4pt]
                (\x,0.50) -- (\x,0.60);

            \node[
                text=graytxt,
                anchor=north
            ] at (\x,0.45) {\lab};
        }

        \draw[densely dashed, gray]
            (4.5,0.7) -- (4.5,6.45);

        \draw[densely dashed, gray]
            (9.0,0.7) -- (9.0,6.45);

        \node[
            text=graytxt,
            anchor=north
        ] at (4.5,0.25)
        {$E[f(x)] = 5.00$};

        \node[
            text=graytxt,
            anchor=south
        ] at (9.0,6.55)
        {$f(x)=5.60$};

        \node[anchor=east] at (2.45,\yA)
        {\textcolor{graytxt}{$S_6$} -- \textbf{Elongated}};

        \node[anchor=east] at (2.45,\yB)
        {\textcolor{graytxt}{$S_2$} -- \textbf{Circle}};

        \node[anchor=east] at (2.45,\yC)
        {\textcolor{graytxt}{$S_4$} -- \textbf{Rectangle}};

        \node[anchor=east] at (2.45,\yD)
        {\textcolor{graytxt}{$S_1$} -- \textbf{Triangle}};

        \node[anchor=east] at (2.45,\yE)
        {\textcolor{graytxt}{$S_5$} -- \textbf{Complex}};

        \fill[poscol]
            (4.50,\yA-0.28)
            --
            (6.45,\yA-0.28)
            --
            (6.60,\yA)
            --
            (6.45,\yA+0.28)
            --
            (4.50,\yA+0.28)
            -- cycle;

        \node[text=white] at (5.45,\yA)
        {$+0.42$};

        \fill[poscol]
            (6.60,\yB-0.28)
            --
            (8.00,\yB-0.28)
            --
            (8.15,\yB)
            --
            (8.00,\yB+0.28)
            --
            (6.60,\yB+0.28)
            -- cycle;

        \node[text=white] at (7.35,\yB)
        {$+0.31$};

        \fill[negcol]
            (8.15,\yC-0.28)
            --
            (7.55,\yC-0.28)
            --
            (7.40,\yC)
            --
            (7.55,\yC+0.28)
            --
            (8.15,\yC+0.28)
            -- cycle;

        \node[text=white] at (7.78,\yC)
        {$-0.15$};

        \fill[poscol]
            (7.40,\yD-0.22)
            --
            (7.68,\yD-0.22)
            --
            (7.80,\yD)
            --
            (7.68,\yD+0.22)
            --
            (7.40,\yD+0.22)
            -- cycle;

        \node[
            text=poscol,
            anchor=west
        ] at (7.90,\yD)
        {$+0.08$};

        \fill[negcol]
            (7.80,\yE-0.22)
            --
            (7.62,\yE-0.22)
            --
            (7.50,\yE)
            --
            (7.62,\yE+0.22)
            --
            (7.80,\yE+0.22)
            -- cycle;

        \node[
            text=negcol,
            anchor=east
        ] at (7.42,\yE)
        {$-0.06$};

    \end{tikzpicture}%
    }

    \caption{
    \textbf{Shape-level explanation produced by \method.}
    In contrast to conventional image attribution methods that associate
    importance with pixels or image regions, each contribution in \method
    corresponds to a morphological primitive characterized by both its
    geometric type and scale.
    Positive Shapley values indicate structures supporting the prediction,
    while negative values indicate opposing evidence.
    For example, the prediction is primarily supported by an
    $S_6$ elongated structure and an $S_2$ circular structure, while an
    $S_4$ rectangular structure provides negative evidence.
    }
    \label{fig:text_expl_shap}

\end{figure}

\paragraph{Shape-level explanation.}
Figure~\ref{fig:text_expl_shap} illustrates how \method summarizes a prediction
through a SHAP-style explanation defined over morphological primitives rather
than over raw pixels.
Each bar in the waterfall plot corresponds to one extracted shape and is
directly associated with three interpretable attributes: its scale level
(e.g., \(S_6\)), its geometric category (e.g., \texttt{Elongated}), and its
signed contribution \(\hat{\phi}_i\).
For instance, the entry
\begin{center}
\texttt{S6 -- Elongated -- $\hat{\phi}=+0.41$}
\end{center}
indicates that a relatively large elongated structure provides positive
evidence for the considered prediction, whereas a negative contribution would
indicate that the corresponding shape opposes it. This representation differs fundamentally from classical image-based
explanations.
Conventional saliency or attribution maps generally indicate \emph{where} the
model looks, but they do not explicitly describe \emph{what kind of visual
structure} supports the decision.

\paragraph{Morphological attribution map.}
\method can also produce classical attribution map since the shape contributions can also be projected back into image space:
\begin{equation}
    H_y(p)
    =
    \sum_{i=1}^{M}
        \hat{\phi}_i
        m_i(p).
    \label{eq:morphological_heatmap}
\end{equation}
Because the Tree of Shapes is hierarchical, a pixel can belong to several
nested structures.
Equation~\eqref{eq:morphological_heatmap} therefore accumulates evidence from
the different morphological scales containing that pixel.

\paragraph{Textual explanation.}

Most importantly, the scale and geometric vocabulary allow the same
attribution to be expressed in words without requiring an external language
model.
For the most influential positive and negative shapes, we generate templates
of the form
\begin{quote}
\emph{``The prediction is mainly supported by an $S_6$ elongated structure
and an $S_2$ circular structure, while an $S_4$ complex shape provides
negative evidence.''}
\end{quote}

Thus, unlike a conventional attribution heatmap that only identifies relevant
locations, \method can describe the morphological nature of the evidence used
by the model.


\subsection{Global Morphological Explanations}
\label{sec:global_explanation}

A key consequence of using a fixed scale and geometric vocabulary is that
explanations from different images can be compared and aggregated.
Pixels from different images have no direct correspondence, and neither do
arbitrary superpixels.
In contrast, an $S_2$ \texttt{Circle} or an $S_6$ \texttt{Elongated}
structure has the same interpretation across samples.

Let $\mathcal{D}_c$ denote the set of images considered when analyzing class
$c$.
For an image $x$, we first normalize its shape contributions as
\begin{equation}
    \widetilde{\phi}_i^{(x)}
    =
    \frac{
        \hat{\phi}_i^{(x)}
    }{
        \sum_j
        \left|
            \hat{\phi}_j^{(x)}
        \right|
        +
        \varepsilon
    }.
    \label{eq:normalized_shap}
\end{equation}

For a geometric category $g$ and scale $s$, we separately aggregate positive
and negative evidence:
\begin{align}
    G_c^{+}(g,s)
    &=
    \frac{1}{|\mathcal{D}_c|}
    \sum_{x\in\mathcal{D}_c}
    \sum_{i=1}^{M_x}
    \mathbbm{1}
    \left[
        g_i=g,\,
        s_i=s
    \right]
    \max
    \left(
        \widetilde{\phi}_i^{(x)},0
    \right),
    \label{eq:global_positive}
    \\
    G_c^{-}(g,s)
    &=
    \frac{1}{|\mathcal{D}_c|}
    \sum_{x\in\mathcal{D}_c}
    \sum_{i=1}^{M_x}
    \mathbbm{1}
    \left[
        g_i=g,\,
        s_i=s
    \right]
    \max
    \left(
        -\widetilde{\phi}_i^{(x)},0
    \right).
    \label{eq:global_negative}
\end{align}

$G_c^{+}$ therefore characterizes the morphological structures that
systematically support class $c$, whereas $G_c^{-}$ characterizes structures
that systematically oppose it.

\section{Experiments}
\label{sec:experiments}

\subsection{Experimental Setup}

\paragraph{Datasets.}
We evaluate MorphoSHAP on five diverse benchmarks spanning natural, astronomical, satellite, biomedical and fine-grained recognition domains:

\begin{itemize}
	\item \textbf{ImageNet}~\cite{deng2009imagenet}: 1,000-class natural image classification. We evaluate on the validation set.
	
	\item \textbf{Galaxy10}~\cite{leung2024galaxy10}: 10-class astronomical object classification (17,736 images). We evaluate on 5,000 stratified test images.
	
	\item \textbf{EuroSAT}~\cite{helber2019eurosat}: 10-class satellite land-use classification (27,000) images. We evaluate on 5,000 stratified test images.
	
	\item \textbf{Waterbirds}~\cite{sagawa2019distributionally}: 2-class bird classification with spurious background correlation (5,794 test images). We evaluate on 5,000 stratified test images.
	
	\item \textbf{TissueMNIST}~\cite{tissuemnist} from MedMNIST~\cite{medmnistv2}: 8-class histology tissue classification (47,280 test images). We evaluate on 5,000 stratified test images.
\end{itemize}

\paragraph{Models.}
We use three standard architectures: ResNet-50~\cite{he2016deep}, ViT-B/16~\cite{dosovitskiy2020image}, and ConvNeXt-Tiny~\cite{liu2022convnet}. ImageNet models use ImageNet-1K pretrained weights; all other models are fine-tuned on their respective training splits. For Grad-CAM family methods, we target the final feature representations of each architecture: the last convolutional block in ResNet (\texttt{layer4[-1]}) and ConvNeXt (\texttt{features[-1][-1]}), and the final layer normalization block in ViT (\texttt{encoder.layers[-1].ln\_1}).

\paragraph{Baselines.}
We compare against three categories of explanation methods:

\begin{itemize}
	\item \textbf{Gradient-based:} Grad-CAM~\cite{selvaraju2017grad}, Grad-CAM++~\cite{chattopadhay2018grad}, Layer-CAM~\cite{jiang2021layercam}, Score-CAM~\cite{wang2020score}, Shapley-CAM~\cite{cai2025cams} and Integrated Gradients (IG)~\cite{sundararajan2017axiomatic}.
	
	\item \textbf{Perturbation-based region methods:} KernelSHAP~\cite{lundberg2017shap} with SLIC superpixels ($K=50$) and KernelSHAP with Otsu binary thresholding, and PartitionSHAP~\cite{lundberg2017unified} (axis-aligned)
	
	\item \textbf{Tree-based Shapley:} ShapBPT~\cite{rashid2026shapbpt} (binary partition tree) mode.
\end{itemize}

\paragraph{Evaluation Metrics.}
Following standard practice in explainability evaluation, we report:
\begin{itemize}
	\item \textbf{Insertion AUC} ($\uparrow$): Area under the curve of predicted-class probability as pixels are inserted from most to least important.
	
	\item \textbf{Deletion AUC} ($\downarrow$): Area under the curve of predicted-class probability as pixels are deleted from most to least important.
	
	\item \textbf{Runtime} (seconds per image, $\downarrow$).
\end{itemize}

Further details are provided in Appendix~\ref{app:eval_metrics}.

\paragraph{Hyperparameters and Ablations.} \method relies on specific hyperparameter configurations to control explanation granularity, which are detailed in Appendix~\ref{Appendix:details}. To evaluate the effects of these structural parameters and our core algorithmic components on the final attributions, we conducted comprehensive ablation studies, the results of which are presented in Appendix~\ref{Appendix:Ablation}.

\begin{table*}[t]
    \centering
    \caption{Quantitative comparison of explanation methods across datasets, averaged across ResNet-50, ViT-B/16, and ConvNeXt-Tiny architectures. \textbf{Bold} indicates best performance. Lower Deletion AUC is better ($\downarrow$); higher Insertion AUC is better ($\uparrow$). Results are Mean $\pm$ Standard Deviation.}
    \label{tab:main_results}
    \resizebox{\textwidth}{!}{%
        \begin{tabular}{l|cc|cc|cc|cc|cc}
            \toprule
            & \multicolumn{2}{c|}{ImageNet} & \multicolumn{2}{c|}{Galaxy10} & \multicolumn{2}{c|}{EuroSAT} & \multicolumn{2}{c|}{Waterbirds} & \multicolumn{2}{c}{TissueMNIST} \\
            Method & Del $\downarrow$ & Ins $\uparrow$ & Del $\downarrow$ & Ins $\uparrow$ & Del $\downarrow$ & Ins $\uparrow$ & Del $\downarrow$ & Ins $\uparrow$ & Del $\downarrow$ & Ins $\uparrow$ \\
            \midrule
            Grad-CAM & 0.253 $\pm$0.180 & 0.685 $\pm$0.179 & 0.258 $\pm$0.292 & 0.555 $\pm$0.302 & 0.453 $\pm$0.242 & 0.555 $\pm$0.228 & 0.420 $\pm$0.289 & 0.852 $\pm$0.184 & 0.174 $\pm$0.192 & 0.460 $\pm$0.239 \\
            Grad-CAM++ & 0.265 $\pm$0.183 & 0.670 $\pm$0.184 & 0.245 $\pm$0.273 & 0.552 $\pm$0.285 & 0.458 $\pm$0.237 & 0.542 $\pm$0.224 & 0.452 $\pm$0.297 & 0.830 $\pm$0.191 & 0.191 $\pm$0.211 & 0.452 $\pm$0.239 \\
            Layer-CAM & 0.259 $\pm$0.181 & 0.674 $\pm$0.184 & 0.170 $\pm$0.198 & 0.604 $\pm$0.244 & 0.459 $\pm$0.236 & 0.550 $\pm$0.221 & 0.451 $\pm$0.296 & 0.832 $\pm$0.191 & 0.189 $\pm$0.228 & 0.394 $\pm$0.259 \\
            Score-CAM & 0.292 $\pm$0.197 & 0.656 $\pm$0.202 & 0.191 $\pm$0.246 & 0.622 $\pm$0.287 & 0.416 $\pm$0.233 & 0.562 $\pm$0.214 & 0.444 $\pm$0.295 & 0.834 $\pm$0.205 & 0.240 $\pm$0.238 & 0.377 $\pm$0.247 \\
            Shapley-CAM & 0.253 $\pm$0.180 & 0.685 $\pm$0.179 & 0.230 $\pm$0.286 & 0.581 $\pm$0.296 & 0.434 $\pm$0.246 & 0.574 $\pm$0.219 & 0.420 $\pm$0.289 & 0.852 $\pm$0.184 & 0.175 $\pm$0.192 & 0.458 $\pm$0.241 \\
            IG & 0.144 $\pm$0.182 & 0.435 $\pm$0.295 & 0.147 $\pm$0.226 & 0.572 $\pm$0.350 & 0.197 $\pm$0.227 & 0.277 $\pm$0.283 & 0.312 $\pm$0.285 & 0.818 $\pm$0.279 & 0.170 $\pm$0.273 & 0.215 $\pm$0.290 \\
            KernelSHAP (SLIC) & 0.223 $\pm$0.163 & 0.736 $\pm$0.173 & 0.138 $\pm$0.154 & 0.776 $\pm$0.200 & 0.364 $\pm$0.219 & 0.616 $\pm$0.258 & 0.337 $\pm$0.280 & 0.890 $\pm$0.208 & 0.157 $\pm$0.218 & 0.445 $\pm$0.292 \\
            KernelSHAP (Otsu) & 0.406 $\pm$0.175 & 0.553 $\pm$0.191 & 0.280 $\pm$0.259 & 0.656 $\pm$0.278 & 0.492 $\pm$0.209 & 0.507 $\pm$0.211 & 0.591 $\pm$0.241 & 0.684 $\pm$0.169 & 0.377 $\pm$0.194 & 0.407 $\pm$0.224 \\
            SHAP-BPT (AA) & 0.280 $\pm$0.217 & 0.847 $\pm$0.126 & 0.083 $\pm$0.109 & 0.843 $\pm$0.153 & 0.384 $\pm$0.197 & 0.683 $\pm$0.195 & 0.280 $\pm$0.268 & 0.889 $\pm$0.214 & 0.143 $\pm$0.190 & 0.552 $\pm$0.245 \\
            SHAP-BPT (BPT) & 0.175 $\pm$0.160 & 0.792 $\pm$0.151 & \textbf{0.074 $\pm$0.117} & 0.811 $\pm$0.197 & 0.330 $\pm$0.215 & 0.633 $\pm$0.256 & 0.292 $\pm$0.273 & \textbf{0.907 $\pm$0.214} & \textbf{0.139 $\pm$0.212} & 0.452 $\pm$0.294 \\
            \midrule
            \textbf{\method (Ours)} & \textbf{0.126 $\pm$0.130} & \textbf{0.876 $\pm$0.081} & 0.116 $\pm$0.086 & \textbf{0.876 $\pm$0.084} & \textbf{0.116 $\pm$0.119} & \textbf{0.885 $\pm$0.151} & \textbf{0.257 $\pm$0.208} & 0.889 $\pm$0.185 & 0.144 $\pm$0.187 & \textbf{0.820 $\pm$0.107} \\
            \bottomrule
        \end{tabular}%
    }
\end{table*}

\begin{table}[t]
    \centering
    \caption{Average runtime per image (seconds) on ImageNet, aggregated across all three model architectures.}
    \label{tab:runtime}
    \resizebox{0.7\columnwidth}{!}{%
    \begin{tabular}{lc}
        \toprule
        Method & Time (s) $\downarrow$ \\
        \midrule
        Grad-CAM & 0.02 \\
        Layer-CAM & 0.02 \\
        Score-CAM & 0.79 \\
        Integrated Gradients (IG) & 1.64 \\
        KernelSHAP (SLIC) & 0.82 \\
        SHAP-BPT (BPT) & 0.29 \\
        \midrule
        \textbf{\method (Ours)} & \textbf{0.13} \\
        \bottomrule
    \end{tabular}
    }
\end{table}

\subsection{Quantitative Comparison}
We report Insertion and Deletion AUC for all methods across the five datasets, averaged across three architectures (ResNet-50, ViT-B/16, and ConvNeXt-Tiny). We also report the average runtime on ImageNet.

\paragraph{Key findings.}
MorphoSHAP achieves the highest Insertion AUC on ImageNet, EuroSAT, Galaxy10, and TissueMNIST, alongside the lowest Deletion AUC on ImageNet, EuroSAT, and Waterbirds. This indicates that our blob-level attributions consistently and accurately capture the regions the model relies on. The performance gap is notably pronounced on EuroSAT and Galaxy10, where objects possess strong macro-morphological structures (e.g., agricultural fields, galactic cores) that our method natively isolates. On datasets like Waterbirds, MorphoSHAP remains highly competitive with state-of-the-art tree-based methods while simultaneously providing the structured textual outputs that region methods fundamentally lack. In Appendix~\ref{Appendix:Ablation} we can see that the results are stable to the change of the hyperparameters. 

\paragraph{Comparison with region-based SHAP.} KernelSHAP with SLIC superpixels performs poorly on structured domains because SLIC boundaries do not respect object contours, diluting attribution. KernelSHAP with Otsu suffers heavily from coarse binary segmentation. ShapBPT provides stronger baselines but still falls short of MorphoSHAP's insertion scores on most datasets and fundamentally lacks the multi-scale shape vocabulary and geometric tagging of our method. 

\paragraph{Computational cost.} As shown in Table~\ref{tab:runtime}, MorphoSHAP is  superior to existing perturbation methods in speed. Averaging 0.13 seconds per image on ImageNet, it is over $6 \times$ faster than SLIC-SHAP (0.82s), over $12\times$ faster than Integrated Gradients (1.64s), and more than $2\times$ faster than ShapBPT (0.29s). Because $M \ll H \cdot W$, MorphoSHAP scales gracefully with image resolution while rivaling the speed of heavier CAM variants.

Detailed comparison of performance metrics are provided in Appendix~\ref{Appendix:Detailed_Quantitative}.

\subsection{Qualitative and Semantic Explanations}
Unlike pixel-level methods that output unstructured heatmaps, MorphoSHAP translates model behavior into a structured semantic vocabulary. By cross-referencing the blob registry $\mathcal{B}$ with the estimated Shapley values, we isolate the specific geometric primitives driving the prediction. 

\begin{figure*}[t]
	\centering
	\includegraphics[width=0.9\linewidth]{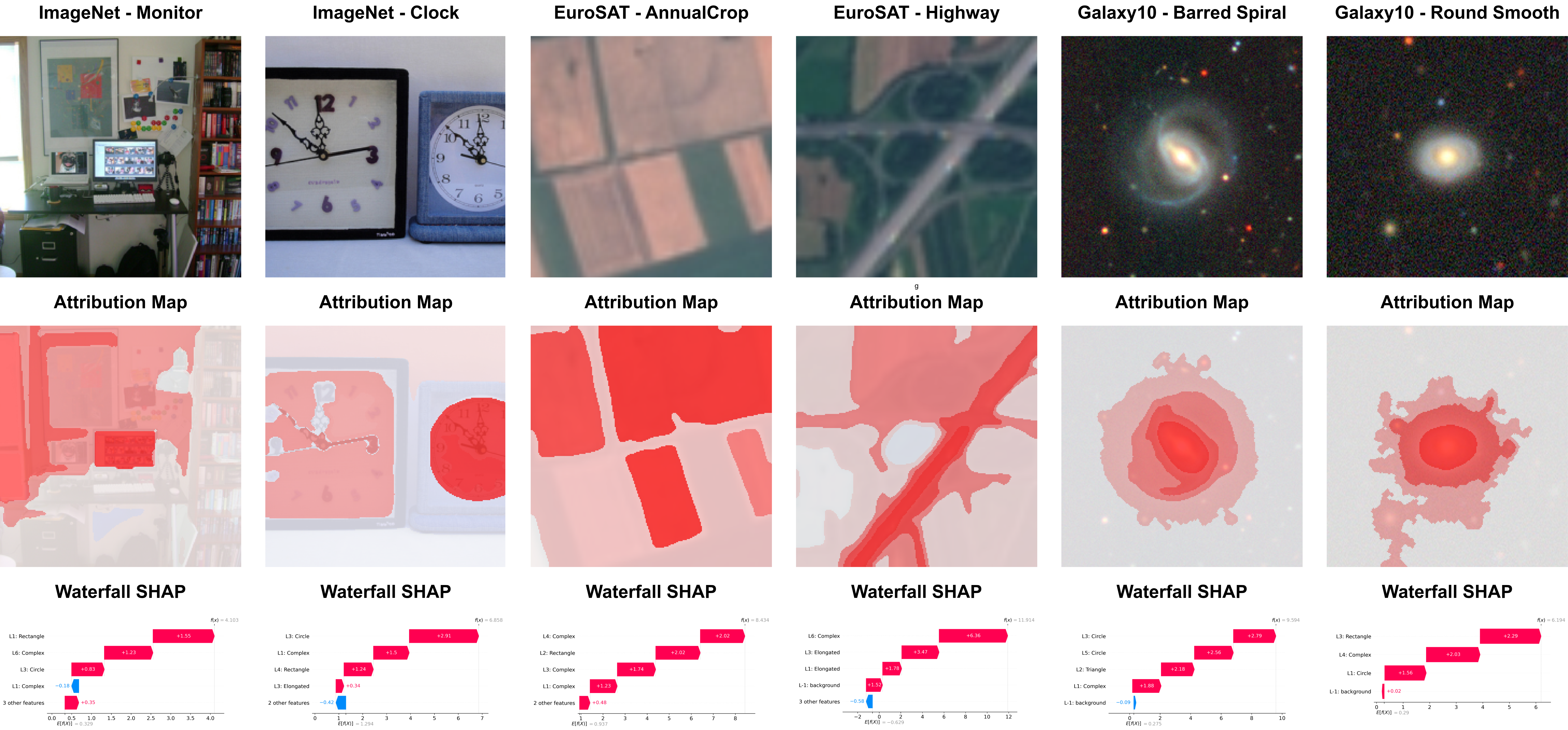}
	\caption{Cross-domain qualitative explanations using MorphoSHAP on ImageNet, EuroSAT, and Galaxy10. The grid displays the original image, the MorphoSHAP attribution heatmap, and a corresponding waterfall plot that isolates specific geometric primitives (e.g., L3: Circle, L4: Rectangle) and quantifies their direct impact on the model's final prediction.}
	\label{fig:cross_domain_qualitative}
\end{figure*}

To demonstrate the versatility of our framework, Figure~\ref{fig:cross_domain_qualitative} visualizes this qualitative analysis across three distinct domains. Because the Tree of Shapes is scale- and contrast-invariant, MorphoSHAP seamlessly adapts to vastly different object morphologies without requiring any domain-specific re-tuning. Unlike standard pixel-level attributions, MorphoSHAP isolates distinct geometric primitives—such as \textit{Circular} clock faces, \textit{Rectangular} street signs or agricultural parcels, and \textit{Complex} galactic structures—and quantifies their exact contribution $f(x)$ to the final prediction via native SHAP waterfall plots. This bridge between pristine visual saliency and structured textual summarization is achieved without relying on external vision-language models, ensuring the explanation remains strictly faithful to the underlying mathematical morphology. Details about the generation of the textual summary are provided in Appendix~\ref{Appendix:Textualexplanation}. One can see in Figure~\ref{fig:cross_domain_qualitative} the two kinds of local explanations provided by MorphoSHAP. Appendix \ref{Appendix:Global_Semantic_Insights} provides the qualitative results of the global explanation.

\subsection{Cross-Domain Global Semantic Insights}
Unlike pixel-attribution methods that are restricted to per-image heatmaps, \method structurally tags each cooperative player prior to attribution, enabling the aggregation of local explanations into dataset-wide global insights. By auditing the primary positive morphological primitives across natural images (ImageNet), overhead satellite imagery (EuroSAT), and astronomical morphology (Galaxy10), we reveal clear, domain-specific geometric dependencies that align closely with physical ground truth. 

For instance, \method shows that predictions for industrial zones rely heavily on \textit{Complex} primitives, while spiral galaxies depend on a dual representation of \textit{Rectangle} central cores and \textit{Complex} arms. Importantly, this vocabulary is modular: additional shape categories can be introduced depending on the structures that are relevant to a given application or domain. A comprehensive breakdown, including our full quantitative distribution table and further analysis of these global morphological dependencies, is provided in Appendix~\ref{Appendix:Global_Semantic_Insights}.

\subsection{User Study}
While automated metrics like Insertion/Deletion AUC measure faithfulness to the model, they do not inherently measure human interpretability. To validate the practical utility of MorphoSHAP, we conducted a user study with 43 participants across 976 independent image evaluation trials using a randomly sampled subset of 150 images spanning 150 distinct ImageNet classes. To ensure non-expert participants could identify the objects, these highly specific classes were augmented with broad, recognizable category labels. The survey was structured into two core tasks: forward simulation (guess-the-class) and subjective preference. 

\paragraph{Forward Simulation.} 
Participants were shown an image masked by the explanation baseline and asked to identify the underlying object class. MorphoSHAP achieved a 89.7\% human guessing accuracy, rivaling the dense pixel-highlighting of Grad-CAM (91.7\%) and heavily outperforming hierarchical region baselines such as SHAP-BPT (85.6\%) and PartitionSHAP (81.6\%). Furthermore, MorphoSHAP imposed the lowest cognitive load on users, requiring a median decision time of only 6.78 seconds per image, the fastest across all evaluated methods. 

\paragraph{Subjective Interpretability.}
When presented with side-by-side attributions and asked to select the most interpretable explanation for a given prediction, MorphoSHAP was overwhelmingly preferred. Participants selected MorphoSHAP as the best explanation in 60.1\% of all trials, vastly exceeding SHAP-BPT (24.6\%) and Grad-CAM (15.2\%). These results demonstrate that extracting discrete, mathematically grounded geometric primitives aligns far better with human visual perception than raw pixel gradients or arbitrary superpixel segmentations.

Detailed information and results from the user study can be found in Appendix~\ref{Appendix:Detailed_User_Study}.
\section{Conclusion}
\label{sec:conclusion}

In this work, we introduced \method, a model-agnostic, post-hoc attribution framework that rethinks the fundamental unit of visual explanation. By leveraging the Tree of Shapes, \method transitions the Shapley game from arbitrary pixel or superpixel grids to meaningful morphological primitives. This shift enables rich, multi-modal explanations that simultaneously capture the spatial location, morphological scale, and geometric category of the visual evidence driving a model's prediction. Our extensive evaluation across diverse domains - including natural, satellite, astronomical and medical imagery - demonstrates that \method not only provides superior insertion and deletion performance compared to existing perturbation methods, but also aligns significantly better with human interpretability.

\paragraph{Limitations and Future Work.}
\method may be less effective when predictions rely mainly on fine textures
rather than well-defined shapes. In addition, the quality of the structured
explanation depends on the tagging function used to describe each shape.
Future work will therefore consider texture-aware decompositions and more
advanced domain-specific or learned tagging strategies.
{
    \small
    \clearpage
    \bibliographystyle{ieeenat_fullname}
    \bibliography{main}
}

\clearpage
\appendix

\renewcommand{\thetable}{S\arabic{table}} 
\renewcommand{\thefigure}{S\arabic{figure}}

\twocolumn[
\begin{center}
    \vspace*{0.3cm}
    {\Large\textbf{
   MorphoSHAP: Rethinking the Unit of Attribution in Explanation for Deep Visual Models
    }}
    
    \vspace{0.2cm}
    
    {\large\textbf{-- Supplementary Material --}}
    
    \vspace{0.7cm}
\end{center}
]

\section{Complementary information on \method}\label{Appendix:Extra_info}

\subsection{Multi-Scale Morphological Characterization}
\label{Appendix:Extra_info_scale}

The Tree of Shapes provides a hierarchy of morphological structures with
different spatial extents. To obtain a simple and shared vocabulary for their
size, we assign each extracted shape $b_i$ to one of eight scale levels.

For a shape $b_i$ with binary mask $m_i$, we define its area as
\begin{equation}
	|b_i|
	=
	\sum_{p\in\Omega} m_i(p),
\end{equation}
and its relative area with respect to the full image as
\begin{equation}
	\rho_i
	=
	\frac{|b_i|}{HW},
	\qquad
	\rho_i\in[0,1],
	\label{eq:relative_area_appendix}
\end{equation}
where $H$ and $W$ denote the image height and width, respectively.
Using a relative area rather than the raw number of pixels makes the scale
definition independent of image resolution.

We consider the ordered scale vocabulary
\begin{equation}
	\mathcal{S}_{\mathrm{scale}}
	=
	\{S_1,S_2,\ldots,S_8\},
\end{equation}
where $S_1$ corresponds to the finest structures and $S_8$ to the coarsest
ones. The scale boundaries are
\begin{equation}
	\boldsymbol{\tau}
	=
	[0,\,
	0.01,\,
	0.05,\,
	0.10,\,
	0.20,\,
	0.30,\,
	0.40,\,
	0.50,\,
	1.00].
\end{equation}
For $k<8$, a shape is assigned to scale $S_k$ according to
\begin{equation}
	s_i=S_k
	\quad\Longleftrightarrow\quad
	\tau_{k-1}\leq\rho_i<\tau_k,
	\label{eq:scale_assignment_appendix}
\end{equation}
while the last interval includes its upper boundary,
$\rho_i\in[0.50,1]$.

Table~\ref{tab:scale_definition} gives the complete scale definition.

\begin{table}[th!]
	\centering
	\caption{
		Definition of the eight morphological scale levels.
		The relative area $\rho_i=|b_i|/(HW)$ measures the fraction of the image
		occupied by shape $b_i$.
	}
	\label{tab:scale_definition}
	\begin{tabular}{c c c}
		\toprule
		\textbf{Scale} &
		\textbf{Relative area $\rho_i$} &
		\textbf{Image area} \\
		\midrule
		$S_1$ & $0 \leq \rho_i < 0.01$    & $<1\%$ \\
		$S_2$ & $0.01 \leq \rho_i < 0.05$ & $1$--$5\%$ \\
		$S_3$ & $0.05 \leq \rho_i < 0.10$ & $5$--$10\%$ \\
		$S_4$ & $0.10 \leq \rho_i < 0.20$ & $10$--$20\%$ \\
		$S_5$ & $0.20 \leq \rho_i < 0.30$ & $20$--$30\%$ \\
		$S_6$ & $0.30 \leq \rho_i < 0.40$ & $30$--$40\%$ \\
		$S_7$ & $0.40 \leq \rho_i < 0.50$ & $40$--$50\%$ \\
		$S_8$ & $0.50 \leq \rho_i \leq 1$ & $\geq50\%$ \\
		\bottomrule
	\end{tabular}
\end{table}

Importantly, these scale levels do not modify the Tree-of-Shapes
decomposition and are not used to generate or filter shapes.
They are assigned \emph{after} shape extraction and serve only as an
interpretable attribute of each morphological primitive.
The scale-enriched representation is therefore
\begin{equation}
	\mathcal{B}_{\mathrm{scale}}(x)
	=
	\left\{
	(b_i,s_i)
	\right\}_{i=1}^{M}.
	\label{eq:shape_scale_appendix}
\end{equation}

This discretization provides a common size vocabulary across images and allows
\method to describe, for example, whether a prediction relies on fine
structures such as an $S_1$ or $S_2$ shape, or on larger structures such as
$S_6$--$S_8$.

\subsection{Rule-based geometric naming}\label{Appendix:Extra_info_naming}

One of the strategies to attribute the label to the shape is entirely deterministic.
It applies a sequence of geometric rules to $d_i$.
The order of the rules is intentional and makes the resulting categories
mutually exclusive.

\begin{enumerate}
	
	\item
	\textbf{Elongated.}
	If
	\begin{equation}
		\operatorname{AR}_i > 2.5,
	\end{equation}
	the shape is labeled \texttt{Elongated}.
	This rule is evaluated first so that strongly anisotropic structures are
	described primarily by their elongation, independently of their precise
	polygonal contour.
	
	\item
	\textbf{Circle.}
	For the remaining shapes, if
	\begin{equation}
		C_i > 0.82,
	\end{equation}
	the shape is labeled \texttt{Circle}.
	The circularity measure reaches $1$ for an ideal circle and decreases as
	the contour departs from circularity.
	
	\item
	\textbf{Triangle, Rectangle, and Polygon.}
	For the remaining shapes, we approximate the contour using the
	Douglas--Peucker algorithm with tolerance
	\begin{equation}
		\epsilon_i=0.04P_i.
	\end{equation}
	Let $n_i$ denote the number of vertices of the simplified contour.
	We assign
	\begin{equation}
		g_i =
		\begin{cases}
			\texttt{Triangle},
			& n_i=3,\\[2mm]
			\texttt{Rectangle},
			& n_i=4,\\[2mm]
			\texttt{Polygon},
			& n_i\in\{5,6\}.
		\end{cases}
		\label{eq:polygon_rules}
	\end{equation}
	
	\item
	\textbf{Complex shape.}
	Every structure that does not satisfy the previous criteria is labeled
	\texttt{Complex}.
	This category captures morphological structures whose geometry cannot be
	faithfully summarized by one of the elementary geometric primitives in
	our vocabulary.
	
\end{enumerate}

We denote this deterministic mapping by
\begin{equation}
	g_i
	=
	h_{\mathrm{rule}}(d_i).
	\label{eq:rule_classifier}
\end{equation}

\subsection{Learned Geometric Naming (MLP Variant)}\label{Appendix:Extra_info_Learned}

While the rule-based approach operates on the compact 6-dimensional descriptor $d_i$, the learned approach requires a richer representation to capture complex structural nuances. For the MLP variant, we extract an expanded 24-dimensional geometric feature vector for each shape contour. 

This expanded feature set includes:
\begin{itemize}
	\item \textbf{Normalized Basic Metrics:} Area and perimeter normalized by the total image area and dimensions.
	\item \textbf{Bounding Geometries:} Ratios of the contour area to its bounding box, minimum enclosing circle, convex hull, and minimum area rectangle.
	\item \textbf{Polygonal and Convexity Features:} Vertex counts at multiple $\epsilon$-tolerances (0.02, 0.05, 0.10) using the Douglas-Peucker algorithm, alongside normalized convexity defect counts and depths.
	\item \textbf{Invariant Moments:} The seven Hu Moments (log-transformed) to capture rotation-, scale-, and translation-invariant shape characteristics.
	\item \textbf{Spatial Distribution:} The standard deviation and maximum of the distances from the contour points to the shape's geometric centroid.
\end{itemize}

Our classifier, \texttt{ShapeMLP}, is parameterized as a Multi-Layer Perceptron (MLP) consisting of three hidden layers with dimensions $(256, 128, 64)$. To ensure stable training across disparate geometric scales, we apply 1D Batch Normalization directly to the input features and after each hidden linear layer. The network utilizes ReLU activations and a dropout rate of $p=0.3$ to prevent overfitting.

\paragraph{Dataset Generation and Training.} 
To train the classifier without requiring exhaustive human annotation, we programmatically generated a diverse synthetic dataset of isolated morphological primitives. Using OpenCV, we rendered over 5,000 shapes with highly randomized geometric parameters - varying scales, aspect ratios, rotation angles, and contour noise - across randomized contrast intensities to accurately emulate the varied output of the Tree of Shapes. The dataset was split into training, validation, and holdout test sets. The network was trained using the Adam optimizer and Cross-Entropy Loss, relying on 1D Batch Normalization and dropout ($p=0.3$) to ensure stable convergence and prevent overfitting. Evaluated on a 500-sample holdout test set, the trained MLP achieved near-perfect accuracy ($\approx 99\%$), demonstrating that the expanded MLP-based descriptor provides a highly separable feature space for identifying our five target morphological categories.

\subsection{Comparisons of the Two Namings}\label{Appendix:Extra_info_comparisons}

The rule-based mapping ($h_{\mathrm{rule}}$) and the learned classifier (\texttt{ShapeMLP}) offer distinctly complementary advantages for morphological attribution.

\textbf{Rule-Based ($h_{\mathrm{rule}}$):}
The primary advantage of the rule-based heuristic is absolute transparency and zero-shot generalization. Because the thresholds rely on explicit, mathematically bounded properties (e.g., circularity and solidity), it seamlessly adapts to vastly different domains---from macroscopic satellite structures to microscopic histology---without requiring any domain-specific training data. Furthermore, it operates strictly on the compact descriptor $d_i$, making it computationally inexpensive.

\textbf{Learned Variant (\texttt{ShapeMLP}):} 
Conversely, the MLP excels at resolving ambiguous edge cases by leveraging the richer 24-dimensional feature space. For instance, real-world object contours frequently suffer from camera perspective skew, discretization noise, and partial occlusion, which easily trip up strict rule-based polygonal approximations. {Instead of depending on basic aspect ratios, the MLP navigates this noise by combining more complex shape features like Hu moments and convexity defects.} However, this method requires a labeled morphological dataset for training and inherently binds the explanation quality to the distribution of that training data.

For the core experiments in the main text, we utilized the Learned Variant (MLP) approach to ensure maximum transparency, domain-independence, and reproducibility across the five diverse benchmark datasets. In Table~\ref{tab:gtsrb_geometry_eval} we compare the performance of the two naming strategies on a real dataset. 

\subsection{Cross-Domain Global Semantic Insights}\label{Appendix:Global_Semantic_Insights}
Unlike pixel-attribution methods that are restricted to per-image heatmaps, MorphoSHAP structurally tags each cooperative player prior to attribution, enabling the aggregation of local explanations into dataset-wide global insights. To demonstrate the cross-domain utility of MorphoSHAP, we audited the primary positive morphological primitives across three distinct vision domains: natural images (\textit{ImageNet}), overhead satellite imagery (\textit{EuroSAT}), and astronomical morphology (\textit{Galaxy10}).

As detailed in Table~\ref{tab:global_insights}, aggregating MorphoSHAP primitives reveals clear, domain-specific geometric dependencies that align closely with physical ground truth:
\begin{itemize}
	\item \textbf{Natural Images (ImageNet):} Man-made object classes are heavily driven by canonical geometric bounds. ResNet-50 relies predominantly on \textit{Rectangle} primitives when identifying \textit{Street Signs} ($46.2\%$) and \textit{Wall Clocks} ($34.3\%$).
	\item \textbf{Satellite Remote Sensing (EuroSAT):} Overhead imagery features complex, multi-facility structures. MorphoSHAP reveals that predictions for \textit{Industrial Zones} are overwhelmingly driven by \textit{Complex Polygon} primitives ($64.0\%$), accurately reflecting the non-convex, sprawling layouts of warehouses and shipping docks.
	\item \textbf{Astronomical Imaging (Galaxy10):} MorphoSHAP successfully disentangles galactic structures. While \textit{Disk Edge-On} galaxies rely heavily on rectilinear central profiles ($76.0\%$ \textit{Rectangle}), \textit{Spiral Galaxies} exhibit a strong dual reliance on central cores ($44.0\%$ \textit{Rectangle}) and sprawling spiral arms ($36.0\%$ \textit{Complex Polygon}).
\end{itemize}

This cross-domain auditing capability demonstrates that MorphoSHAP captures authentic, domain-appropriate structural representations, revealing the underlying geometric priors learned by the neural network. Furthermore, while the insights presented in Table~\ref{tab:global_insights} were generated using our lightweight, rule-based geometric heuristic, the MorphoSHAP framework is inherently extensible. The vocabulary of the shapes exposes a modular tagging interface allowing practitioners to seamlessly integrate custom tagging functions. By defining any function that maps an isolated binary morphological blob to a semantic string, researchers can deploy sophisticated domain-specific taggers—including specialized deep neural networks or pre-trained vision-language models—to extract arbitrarily complex semantic explanations tailored to their specific use case.

\begin{table}[h]
	\centering
	\resizebox{\linewidth}{!}{%
		\begin{tabular}{llccccc}
			\toprule
			\textbf{Dataset} & \textbf{Class} & \textbf{Triangle} & \textbf{Circle} & \textbf{Rectangle} & \textbf{Elongated} & \textbf{Complex} \\
			\midrule
			\multirow{4}{*}{ImageNet}
			& Wall Clock & 0.0\% & \textbf{32.0\%} & 26.0\% & 22.0\% & 20.0\% \\
			& Monitor & 0.0\% & 30.0\% & \textbf{34.0\%} & 14.0\% & 22.0\% \\
			& Street Sign & 2.0\% & 20.0\% & 10.0\% & 32.0\% & \textbf{36.0\%} \\
			& Traffic Light & 0.0\% & 18.4\% & 10.2\% & \textbf{40.8\%} & 30.6\% \\
			\midrule
			\multirow{6}{*}{EuroSAT}
			& AnnualCrop & 4.0\% & 16.0\% & 10.0\% & \textbf{40.0\%} & 30.0\% \\
			& Highway & 0.0\% & 0.0\% & 6.0\% & 24.0\% & \textbf{70.0\%} \\
			& Industrial & 0.0\% & 10.0\% & 14.0\% & 8.0\% & \textbf{68.0\%} \\
			& Residential & 0.0\% & 0.0\% & 8.0\% & 12.0\% & \textbf{80.0\%} \\
			& River & 4.0\% & 6.0\% & 20.0\% & \textbf{50.0\%} & 20.0\% \\
			& SeaLake & 0.0\% & 8.1\% & 18.9\% & 16.2\% & \textbf{56.8\%} \\
			\midrule
			\multirow{5}{*}{Galaxy10}
			& Merging Galaxies & 0.0\% & \textbf{48.0\%} & 16.0\% & 18.0\% & 18.0\% \\
			& Round Smooth Galaxies & 0.0\% & \textbf{64.0\%} & 24.0\% & 2.0\% & 10.0\% \\
			& In-between Round Smooth Galaxies & 0.0\% & 22.0\% & \textbf{36.0\%} & 12.0\% & 30.0\% \\
			& Barred Spiral Galaxies & 0.0\% & \textbf{66.0\%} & 16.0\% & 6.0\% & 12.0\% \\
			& Unbarred Tight Spiral Galaxies & 0.0\% & \textbf{70.0\%} & 14.0\% & 10.0\% & 6.0\% \\
			\bottomrule
		\end{tabular}%
	}
	\caption{Cross-Domain Global Semantic Aggregation. Distribution of the highest-attributed MorphoSHAP primitives across natural, satellite, and astronomical datasets. The data aggregates shape counts across all 8 scale layers.}
	\label{tab:global_insights}
\end{table}

\subsection{Textual explanation}
\label{Appendix:Textualexplanation}

\method translates spatial attributions into structured textual summaries. This is achieved deterministically through a rule-based natural language generation pipeline rather than relying on external Vision-Language Models (VLMs) or Large Language Models (LLMs). This guarantees that the generated explanation remains strictly faithful to the underlying Shapley values and avoids linguistic hallucinations.

\paragraph{Generation Pipeline.}
Given the blob registry $\mathcal{B}$ and estimated Shapley values $\hat{\phi}$, the text generation module operates in five sequential steps:

\begin{enumerate}
    \item \textbf{Aggregation and Area Tracking:} We ignore background blobs (where $g_i = \text{background}$). For each active morphological shape category $g \in \mathcal{G}$, we sum the total Shapley contributions $\Phi(g) = \sum_{b_i \in g} \hat{\phi}_i$. Simultaneously, we track the maximum area ratio $\rho_{\max}(g) = \max_{b_i \in g} \frac{|b_i|}{HW}$ occupied by the largest blob of that shape category.
    
    \item \textbf{Filtering and Ranking:} We filter for shape categories that actively support the target class prediction ($\Phi(g) > 0$). The surviving categories are sorted in descending order by their aggregated Shapley contribution, and we select up to the top 3 dominant shape contributors.
    
    \item \textbf{Size Adjective Mapping:} To describe spatial scale, the maximum relative area ratio $\rho_{\max}(g)$ of each top contributor is mapped to a discrete size adjective:
    \begin{equation}
    \text{Size}(g) = 
    \begin{cases} 
        \text{large}, & \text{if } \rho_{\max}(g) > 0.15, \\
        \text{medium}, & \text{if } 0.05 < \rho_{\max}(g) \le 0.15, \\
        \text{small}, & \text{if } \rho_{\max}(g) \le 0.05.
    \end{cases}
    \end{equation}

    \item \textbf{Geometry Adjective Conversion:} Discrete geometric labels are converted into descriptive adjectives (e.g., \textit{Circle} $\rightarrow$ \textit{circular}, \textit{Rectangle} $\rightarrow$ \textit{rectangular}, \textit{Triangle} $\rightarrow$ \textit{triangular}, \textit{Elongated} $\rightarrow$ \textit{elongated}, \textit{Complex} $\rightarrow$ \textit{complex}). Each contributor is formatted into a noun phrase of the form: \textit{``[size] [geometry] structure''}.

    \item \textbf{Grammatical Assembly:} The extracted phrases are injected into deterministic template strings based on the number of top positive contributors ($K \in \{0, 1, 2, 3\}$):
    \begin{itemize}
        \item $K = 0$: \textit{The prediction is primarily supported by the background context.}
        \item $K = 1$: \textit{The prediction is mainly supported by a [Phrase$_1$].}
        \item $K = 2$: \textit{The prediction is mainly supported by a [Phrase$_1$] and a [Phrase$_2$].}
        \item $K = 3$: \textit{The prediction is mainly supported by a [Phrase$_1$], a [Phrase$_2$], and a [Phrase$_3$].}
    \end{itemize}
\end{enumerate}

For example, if the top three positive shape categories for a prediction are a large complex blob ($\Phi = 0.42, \rho_{\max} = 0.22$), a medium rectangular blob ($\Phi = 0.18, \rho_{\max} = 0.08$), and a small circular blob ($\Phi = 0.09, \rho_{\max} = 0.02$), the pipeline outputs: 
\begin{quote}
    \textit{``The prediction is mainly supported by a large complex structure, a medium rectangular structure, and a small circular structure.''}
\end{quote}

\section{Details on the Evaluation Metrics}
\label{app:eval_metrics}

We evaluate explanation faithfulness using the standard
\emph{Insertion} and \emph{Deletion} metrics, together with the runtime required
to generate an explanation.

\paragraph{Insertion AUC.}
Insertion measures whether the regions identified as important are sufficient
to recover the model prediction.
Starting from a reference image $x_0$, explanatory units are progressively
restored from the most to the least important.
Let $\pi=(\pi_1,\ldots,\pi_M)$ denote the ranking of the $M$ explanatory units.
After inserting the first $k$ units, we evaluate the predicted-class score
\begin{equation}
    I(k)
    =
    f_y\!\left(x^{\mathrm{ins}}_k\right),
    \qquad k=0,\ldots,M.
\end{equation}
The Insertion AUC is the area under the curve
$f_y(x^{\mathrm{ins}}_k)$ as increasingly more evidence is added.
A \emph{higher} AUC indicates that the most important units identified by the
explanation rapidly recover the model prediction.

\paragraph{Deletion AUC.}
Deletion follows the opposite procedure and measures whether removing the most
important regions rapidly decreases the prediction.
Starting from the original image $x$, explanatory units are removed according
to the same importance ranking:
\begin{equation}
    D(k)
    =
    f_y\!\left(x^{\mathrm{del}}_k\right),
    \qquad k=0,\ldots,M.
\end{equation}
A faithful explanation should identify regions whose removal strongly affects
the prediction; therefore, a \emph{lower} Deletion AUC is better.

\paragraph{MorphoSHAP evaluation.}
For standard pixel- or region-based methods, insertion and deletion operate on
their corresponding explanatory units.
For \method, the units are the morphological shapes
$\mathcal{B}(x)=\{b_1,\ldots,b_M\}$ rather than individual pixels.
The curves are therefore constructed by progressively inserting or deleting
morphological blobs according to their attribution ranking.
When a shape is absent, its pixels are replaced by the same reference value
$x_0$ used in the Shapley perturbation process.

This shape-level evaluation is consistent with the main goal of \method:
faithfulness is measured in the same morphological space in which the
explanation is defined.

\paragraph{Runtime.}
We additionally report the average runtime required to produce an explanation,
in seconds per image.
Lower runtime indicates a more computationally efficient explanation method.

\paragraph{Sampling budget.}
For a fair comparison between perturbation-based approaches, we use the same
KernelSHAP sampling budget of
\begin{equation}
    N=1024
\end{equation}
coalitions for all such methods.
\section{Experiments}

\subsection{Detailed Quantitative Results per Architecture}
\label{Appendix:Detailed_Quantitative}

In the main text (Table~\ref{tab:main_results}), we report the Insertion and Deletion AUC metrics averaged across all three evaluated architectures (ResNet-50, ViT-B/16, and ConvNeXt-Tiny) to provide a holistic view of explanation faithfulness. 

In this section, we provide the detailed, per-architecture breakdowns. Tables \ref{tab:resnet_results}, \ref{tab:vit_results}, and \ref{tab:convnext_results} report the individual performance on ResNet-50, ViT-B/16, and ConvNeXt-Tiny, respectively. These unaggregated results demonstrate that \method maintains consistent faithfulness across structurally diverse model families, from standard convolutions to patch-based vision transformers.

\begin{table*}[t]
    \centering
    \caption{Quantitative comparison of explanation methods on ResNet-50. Lower Deletion AUC is better ($\downarrow$); higher Insertion AUC is better ($\uparrow$). \textbf{Bold} indicates best performance.}
    \label{tab:resnet_results}
    \resizebox{\textwidth}{!}{%
        \begin{tabular}{l|cc|cc|cc|cc|cc}
            \toprule
            & \multicolumn{2}{c|}{ImageNet} & \multicolumn{2}{c|}{Galaxy10} & \multicolumn{2}{c|}{EuroSAT} & \multicolumn{2}{c|}{Waterbirds} & \multicolumn{2}{c}{TissueMNIST} \\
            Method & Del $\downarrow$ & Ins $\uparrow$ & Del $\downarrow$ & Ins $\uparrow$ & Del $\downarrow$ & Ins $\uparrow$ & Del $\downarrow$ & Ins $\uparrow$ & Del $\downarrow$ & Ins $\uparrow$ \\
            \midrule
            Grad-CAM & 0.260 $\pm$0.184 & 0.705 $\pm$0.173 & 0.147 $\pm$0.183 & 0.668 $\pm$0.190 & 0.354 $\pm$0.267 & 0.552 $\pm$0.265 & 0.412 $\pm$0.277 & 0.876 $\pm$0.158 & 0.173 $\pm$0.241 & 0.371 $\pm$0.280 \\
            Grad-CAM++ & 0.271 $\pm$0.187 & 0.690 $\pm$0.180 & 0.149 $\pm$0.184 & 0.664 $\pm$0.190 & 0.355 $\pm$0.270 & 0.549 $\pm$0.267 & 0.428 $\pm$0.279 & 0.862 $\pm$0.154 & 0.194 $\pm$0.272 & 0.349 $\pm$0.281 \\
            Layer-CAM & 0.267 $\pm$0.186 & 0.695 $\pm$0.179 & 0.150 $\pm$0.185 & 0.662 $\pm$0.192 & 0.358 $\pm$0.269 & 0.546 $\pm$0.268 & 0.432 $\pm$0.279 & 0.861 $\pm$0.153 & 0.195 $\pm$0.274 & 0.341 $\pm$0.283 \\
            Score-CAM & 0.283 $\pm$0.194 & 0.694 $\pm$0.182 & 0.148 $\pm$0.189 & 0.662 $\pm$0.198 & 0.357 $\pm$0.265 & 0.549 $\pm$0.276 & 0.425 $\pm$0.275 & 0.864 $\pm$0.151 & 0.206 $\pm$0.277 & 0.314 $\pm$0.285 \\
            Shapley-CAM & 0.260 $\pm$0.184 & 0.705 $\pm$0.173 & 0.147 $\pm$0.183 & 0.668 $\pm$0.190 & 0.354 $\pm$0.267 & 0.552 $\pm$0.265 & 0.412 $\pm$0.277 & 0.876 $\pm$0.158 & 0.173 $\pm$0.241 & 0.371 $\pm$0.280 \\
            IG & 0.142 $\pm$0.183 & 0.397 $\pm$0.287 & 0.194 $\pm$0.267 & 0.639 $\pm$0.353 & 0.138 $\pm$0.243 & 0.138 $\pm$0.243 & 0.260 $\pm$0.252 & 0.640 $\pm$0.276 & 0.187 $\pm$0.355 & 0.187 $\pm$0.355 \\
            KernelSHAP (SLIC) & 0.227 $\pm$0.165 & 0.749 $\pm$0.171 & 0.181 $\pm$0.182 & 0.700 $\pm$0.208 & 0.297 $\pm$0.252 & 0.478 $\pm$0.307 & 0.304 $\pm$0.257 & 0.902 $\pm$0.171 & 0.172 $\pm$0.263 & 0.307 $\pm$0.295 \\
            KernelSHAP (Otsu) & 0.416 $\pm$0.179 & 0.572 $\pm$0.188 & 0.320 $\pm$0.244 & 0.527 $\pm$0.266 & 0.475 $\pm$0.231 & 0.430 $\pm$0.239 & 0.538 $\pm$0.244 & 0.673 $\pm$0.178 & 0.378 $\pm$0.188 & 0.350 $\pm$0.237 \\
            SHAP-BPT (AA) & 0.301 $\pm$0.226 & 0.869 $\pm$0.110 & 0.095 $\pm$0.130 & 0.796 $\pm$0.179 & 0.346 $\pm$0.215 & 0.661 $\pm$0.240 & 0.271 $\pm$0.252 & \textbf{0.926 $\pm$0.169} & \textbf{0.140 $\pm$0.184} & 0.503 $\pm$0.280 \\
            SHAP-BPT (BPT) & 0.182 $\pm$0.166 & 0.802 $\pm$0.146 & \textbf{0.091 $\pm$0.141} & 0.760 $\pm$0.207 & 0.273 $\pm$0.239 & 0.530 $\pm$0.293 & 0.259 $\pm$0.239 & 0.911 $\pm$0.176 & 0.142 $\pm$0.243 & 0.298 $\pm$0.304 \\
            \midrule
            \textbf{\method (Ours)} & \textbf{0.131 $\pm$0.137} & \textbf{0.881 $\pm$0.080} & 0.137 $\pm$0.108 & \textbf{0.851 $\pm$0.112} & \textbf{0.119 $\pm$0.125} & \textbf{0.864 $\pm$0.166} & \textbf{0.224 $\pm$0.185} & 0.897 $\pm$0.163 & 0.146 $\pm$0.239 & \textbf{0.873 $\pm$0.053} \\
            \bottomrule
        \end{tabular}%
    }
\end{table*}

\begin{table*}[t]
    \centering
    \caption{Quantitative comparison of explanation methods on ViT-B/16. Lower Deletion AUC is better ($\downarrow$); higher Insertion AUC is better ($\uparrow$). \textbf{Bold} indicates best performance.}
    \label{tab:vit_results}
    \resizebox{\textwidth}{!}{%
        \begin{tabular}{l|cc|cc|cc|cc|cc}
            \toprule
            & \multicolumn{2}{c|}{ImageNet} & \multicolumn{2}{c|}{Galaxy10} & \multicolumn{2}{c|}{EuroSAT} & \multicolumn{2}{c|}{Waterbirds} & \multicolumn{2}{c}{TissueMNIST} \\
            Method & Del $\downarrow$ & Ins $\uparrow$ & Del $\downarrow$ & Ins $\uparrow$ & Del $\downarrow$ & Ins $\uparrow$ & Del $\downarrow$ & Ins $\uparrow$ & Del $\downarrow$ & Ins $\uparrow$ \\
            \midrule
            Grad-CAM & 0.408 $\pm$0.252 & 0.416 $\pm$0.238 & 0.495 $\pm$0.334 & 0.319 $\pm$0.309 & 0.459 $\pm$0.220 & 0.441 $\pm$0.213 & 0.360 $\pm$0.277 & \textbf{0.930 $\pm$0.108} & 0.176 $\pm$0.183 & 0.477 $\pm$0.216 \\
            Grad-CAM++ & 0.427 $\pm$0.229 & 0.424 $\pm$0.202 & 0.414 $\pm$0.326 & 0.386 $\pm$0.306 & 0.481 $\pm$0.213 & 0.430 $\pm$0.197 & 0.387 $\pm$0.257 & 0.892 $\pm$0.157 & 0.197 $\pm$0.196 & 0.488 $\pm$0.205 \\
            Layer-CAM & 0.405 $\pm$0.224 & 0.400 $\pm$0.183 & 0.208 $\pm$0.215 & 0.525 $\pm$0.248 & 0.489 $\pm$0.213 & 0.438 $\pm$0.186 & 0.539 $\pm$0.332 & 0.866 $\pm$0.180 & 0.190 $\pm$0.246 & 0.323 $\pm$0.254 \\
            Score-CAM & 0.312 $\pm$0.229 & 0.511 $\pm$0.243 & 0.243 $\pm$0.292 & 0.546 $\pm$0.331 & 0.373 $\pm$0.211 & 0.526 $\pm$0.182 & 0.467 $\pm$0.317 & 0.880 $\pm$0.175 & 0.247 $\pm$0.241 & 0.386 $\pm$0.234 \\
            Shapley-CAM & 0.440 $\pm$0.275 & 0.369 $\pm$0.211 & 0.411 $\pm$0.371 & 0.397 $\pm$0.343 & 0.401 $\pm$0.229 & 0.500 $\pm$0.211 & 0.409 $\pm$0.313 & 0.917 $\pm$0.119 & 0.178 $\pm$0.184 & 0.470 $\pm$0.222 \\
            IG & \textbf{0.137 $\pm$0.208} & 0.654 $\pm$0.286 & 0.083 $\pm$0.131 & 0.474 $\pm$0.310 & 0.219 $\pm$0.233 & 0.400 $\pm$0.319 & \textbf{0.143 $\pm$0.281} & 0.907 $\pm$0.244 & \textbf{0.142 $\pm$0.205} & 0.212 $\pm$0.241 \\
            KernelSHAP (SLIC) & 0.291 $\pm$0.231 & 0.818 $\pm$0.156 & 0.117 $\pm$0.136 & 0.809 $\pm$0.188 & 0.350 $\pm$0.179 & 0.598 $\pm$0.206 & 0.386 $\pm$0.299 & 0.882 $\pm$0.230 & 0.169 $\pm$0.241 & 0.429 $\pm$0.271 \\
            KernelSHAP (Otsu) & 0.476 $\pm$0.209 & 0.631 $\pm$0.193 & 0.229 $\pm$0.248 & 0.696 $\pm$0.265 & 0.430 $\pm$0.178 & 0.504 $\pm$0.206 & 0.647 $\pm$0.233 & 0.685 $\pm$0.167 & 0.361 $\pm$0.200 & 0.418 $\pm$0.225 \\
            SHAP-BPT (AA) & 0.244 $\pm$0.215 & 0.771 $\pm$0.140 & \textbf{0.061 $\pm$0.078} & 0.862 $\pm$0.163 & 0.385 $\pm$0.201 & 0.649 $\pm$0.201 & 0.312 $\pm$0.291 & 0.895 $\pm$0.235 & 0.176 $\pm$0.250 & 0.482 $\pm$0.233 \\
            SHAP-BPT (BPT) & 0.215 $\pm$0.208 & 0.866 $\pm$0.133 & 0.067 $\pm$0.114 & 0.833 $\pm$0.191 & 0.309 $\pm$0.190 & 0.573 $\pm$0.223 & 0.343 $\pm$0.300 & 0.904 $\pm$0.240 & 0.152 $\pm$0.240 & 0.443 $\pm$0.254 \\
            \midrule
            \textbf{\method (Ours)} & 0.165 $\pm$0.159 & \textbf{0.892 $\pm$0.113} & 0.106 $\pm$0.078 & \textbf{0.889 $\pm$0.062} & \textbf{0.102 $\pm$0.125} & \textbf{0.872 $\pm$0.179} & 0.296 $\pm$0.220 & 0.882 $\pm$0.200 & 0.162 $\pm$0.200 & \textbf{0.778 $\pm$0.124} \\
            \bottomrule
        \end{tabular}%
    }
\end{table*}

\begin{table*}[t]
    \centering
    \caption{Quantitative comparison of explanation methods on ConvNeXt-Tiny. Lower Deletion AUC is better ($\downarrow$); higher Insertion AUC is better ($\uparrow$). \textbf{Bold} indicates best performance.}
    \label{tab:convnext_results}
    \resizebox{\textwidth}{!}{%
        \begin{tabular}{l|cc|cc|cc|cc|cc}
            \toprule
            & \multicolumn{2}{c|}{ImageNet} & \multicolumn{2}{c|}{Galaxy10} & \multicolumn{2}{c|}{EuroSAT} & \multicolumn{2}{c|}{Waterbirds} & \multicolumn{2}{c}{TissueMNIST} \\
            Method & Del $\downarrow$ & Ins $\uparrow$ & Del $\downarrow$ & Ins $\uparrow$ & Del $\downarrow$ & Ins $\uparrow$ & Del $\downarrow$ & Ins $\uparrow$ & Del $\downarrow$ & Ins $\uparrow$ \\
            \midrule
            Grad-CAM & 0.218 $\pm$0.152 & 0.583 $\pm$0.172 & 0.132 $\pm$0.160 & 0.678 $\pm$0.243 & 0.546 $\pm$0.194 & 0.672 $\pm$0.118 & 0.427 $\pm$0.301 & 0.828 $\pm$0.205 & 0.174 $\pm$0.139 & 0.532 $\pm$0.181 \\
            Grad-CAM++ & 0.233 $\pm$0.159 & 0.567 $\pm$0.169 & 0.173 $\pm$0.202 & 0.607 $\pm$0.266 & 0.538 $\pm$0.180 & 0.647 $\pm$0.128 & 0.475 $\pm$0.312 & 0.798 $\pm$0.218 & 0.181 $\pm$0.144 & 0.519 $\pm$0.183 \\
            Layer-CAM & 0.216 $\pm$0.149 & 0.571 $\pm$0.176 & 0.153 $\pm$0.185 & 0.624 $\pm$0.265 & 0.530 $\pm$0.181 & 0.666 $\pm$0.119 & 0.470 $\pm$0.310 & 0.803 $\pm$0.218 & 0.181 $\pm$0.143 & 0.517 $\pm$0.185 \\
            Score-CAM & 0.336 $\pm$0.206 & 0.463 $\pm$0.192 & 0.182 $\pm$0.236 & 0.658 $\pm$0.301 & 0.519 $\pm$0.179 & 0.611 $\pm$0.156 & 0.464 $\pm$0.312 & 0.803 $\pm$0.243 & 0.266 $\pm$0.181 & 0.431 $\pm$0.200 \\
            Shapley-CAM & 0.218 $\pm$0.152 & 0.583 $\pm$0.172 & 0.132 $\pm$0.160 & 0.678 $\pm$0.243 & 0.546 $\pm$0.194 & 0.672 $\pm$0.118 & 0.427 $\pm$0.301 & 0.828 $\pm$0.205 & 0.174 $\pm$0.139 & 0.532 $\pm$0.181 \\
            IG & 0.156 $\pm$0.176 & 0.607 $\pm$0.266 & 0.164 $\pm$0.240 & 0.602 $\pm$0.364 & 0.234 $\pm$0.188 & 0.292 $\pm$0.212 & 0.269 $\pm$0.423 & \textbf{0.908 $\pm$0.224} & 0.181 $\pm$0.232 & 0.246 $\pm$0.257 \\
            KernelSHAP (SLIC) & 0.195 $\pm$0.133 & 0.663 $\pm$0.169 & 0.117 $\pm$0.130 & 0.820 $\pm$0.180 & 0.444 $\pm$0.194 & 0.771 $\pm$0.140 & 0.320 $\pm$0.275 & 0.887 $\pm$0.218 & 0.131 $\pm$0.119 & 0.598 $\pm$0.231 \\
            KernelSHAP (Otsu) & 0.348 $\pm$0.130 & 0.453 $\pm$0.172 & 0.290 $\pm$0.275 & 0.745 $\pm$0.255 & 0.571 $\pm$0.189 & 0.589 $\pm$0.145 & 0.586 $\pm$0.235 & 0.693 $\pm$0.160 & 0.392 $\pm$0.191 & 0.453 $\pm$0.196 \\
            SHAP-BPT (AA) & 0.179 $\pm$0.120 & 0.741 $\pm$0.145 & 0.094 $\pm$0.110 & 0.872 $\pm$0.092 & 0.422 $\pm$0.163 & 0.740 $\pm$0.107 & 0.258 $\pm$0.255 & 0.846 $\pm$0.223 & \textbf{0.112 $\pm$0.097} & 0.672 $\pm$0.159 \\
            SHAP-BPT (BPT) & 0.131 $\pm$0.108 & 0.733 $\pm$0.158 & \textbf{0.065 $\pm$0.089} & 0.839 $\pm$0.183 & 0.406 $\pm$0.190 & 0.796 $\pm$0.140 & 0.275 $\pm$0.270 & 0.906 $\pm$0.222 & 0.122 $\pm$0.134 & 0.615 $\pm$0.228 \\
            \midrule
            \textbf{\method (Ours)} & \textbf{0.097 $\pm$0.071} & \textbf{0.845 $\pm$0.076} & 0.105 $\pm$0.061 & \textbf{0.888 $\pm$0.062} & \textbf{0.127 $\pm$0.102} & \textbf{0.921 $\pm$0.084} & \textbf{0.252 $\pm$0.211} & 0.888 $\pm$0.190 & 0.124 $\pm$0.088 & \textbf{0.807 $\pm$0.107} \\
            \bottomrule
        \end{tabular}%
    }
\end{table*}

\paragraph{Computational Cost per Architecture.}
In addition to the insertion and deletion metrics, we provide a detailed breakdown of the computational cost across different model architectures. Table~\ref{tab:runtime_detailed} reports the average runtime (in seconds per image) on the ImageNet validation set for ResNet-50, ViT-B/16, and ConvNeXt-Tiny. \method consistently remains significantly faster than both standard perturbation approaches (KernelSHAP) and tree-based Shapley estimators (SHAP-BPT) across all structural families, scaling efficiently even on computationally heavy Vision Transformers.

\begin{table*}[h]
    \centering
    \caption{Detailed runtime breakdown (seconds/image) on ImageNet for each architecture. Lower is better ($\downarrow$).}
    \label{tab:runtime_detailed}
    \begin{tabular}{lccc}
        \toprule
        \textbf{Method} & \textbf{ResNet-50} & \textbf{ViT-B/16} & \textbf{ConvNeXt-Tiny} \\
        \midrule
        Grad-CAM & 0.01 & 0.02 & 0.02 \\
        Layer-CAM & 0.01 & 0.02 & 0.02 \\
        Score-CAM & 0.74 & 1.24 & 0.40 \\
        Integrated Gradients (IG) & 0.81 & 3.05 & 1.08 \\
        KernelSHAP (SLIC) & 0.42 & 1.35 & 0.70 \\
        SHAP-BPT (BPT) & 0.24 & 0.34 & 0.28 \\
        \midrule
        \textbf{\method (Ours)} & \textbf{0.09} & \textbf{0.16} & \textbf{0.13} \\
        \bottomrule
    \end{tabular}
\end{table*}

\subsection{Algorithm details}\label{Appendix:details}

The performance and semantic granularity of the \method framework are primarily governed by four core hyperparameters:
\begin{itemize}
    \item \textbf{Scale Thresholds} (\texttt{threshold\_percents}): This dictates the multi-scale granularity of the Tree of Shapes decomposition. As defined in Appendix~\ref{Appendix:Extra_info_scale}, we utilize eight relative area thresholds ($S_1$ through $S_8$) to extract shapes ranging from microscopic (0-1\% of image area) to global ($>$50\% of image area). Changing these boundaries directly alters the size and number of the morphological structures evaluated by the model.

    \item \textbf{Geometric Tagging Function} (\texttt{tagging\_fn}): This dictates how visual semantics are assigned to the extracted blobs. The framework supports switching between a deterministic, Rule-Based heuristic (relying on strict geometric bounds like circularity and aspect ratio) and a Learned MLP (which evaluates a 24-dimension feature vector).

    \item \textbf{Overlap Threshold:} Because the Tree of Shapes is hierarchical, it inherently produces nested and overlapping structures. To constrain the number of Shapley players and prevent redundant attributions, candidate shapes are sorted by area, and any shape that overlaps with previously registered blobs by more than $50\%$ is discarded.

    \item \textbf{Shapley Sampling Budget} (\texttt{nsamples}): When computing the morphological attributions via \texttt{shap.KernelExplainer}, we constrain the number of sampled coalitions. For our main experiments, the budget is set to $N = 1024$ to balance attribution stability with computational efficiency.
\end{itemize}

\subsection{Ablation study}\label{Appendix:Ablation}
To systematically isolate the impact of our design choices on explanation faithfulness and computational efficiency, we conducted an ablation study across a randomly sampled subset of 200 images (100 from ImageNet, 100 from EuroSAT). These were evalauted across all three base architectures (ResNet-50, ViT-B/16 and ConvNeXt-Tiny). We measured Insertion and Deletion AUC alongside runtime. Note that the reported runtimes in this section include the heavy computational overload needed to compute the AUC curves, making them higher than the pure explanation generation times reported in the main text.

\paragraph{Shapley Sampling Budget}
To evaluate the convergence of the KernelSHAP estimator, we swept the sampling budget $N$ from 32 to 4096 coalitions. As shown in Table~\ref{tab:ablation_budget}, faithfulness metrics improve rapidly at lower budgets. However, both Insertion and Deletion AUC effectively plateau after 1024 samples, which the computational runtimes continues to scale. This mathematically validates our baseline choice of $N = 1024$ as the optimal balance between attribution stability and speed.

\paragraph{Morphological Overlap Threshold}
The hierarchical nature of the Tree of Shapes naturally produces overlapping nested structures. We ablated our overlap rejection threshold from 10\% to 90\% (Table~\ref{tab:ablation_overlap}) to test player redundancy. Stricter thresholds (e.g. 10\%) aggressively prune players, resulting in fast computation (1.86s) but heavily penalizing Deletion AUC by discarding critical structural boundaries. Conversely, highly permissive thresholds (e.g. 90\%) marginarrly improve Insertion AUC but cause the player count and runtime to explode (7.59s). The 50\% baseline ensures distinct morphological players while maintaining a highly competitive runtime.

\paragraph{Scale Granularity}
Finally, we compare our baseline 8-level scale definition against uniformly spaced 4-bin, 10-bin and 15-bin configurations, as well as an 8-bin logarithmically spaced distribution (Table~\ref{tab:ablation_scale}). Coarse granularity (4 bins) struggles to isolate specific visual evidence, resulting in a severe drop in Insertion AUC. While a log-spaced distribution mathematically maximized faithfulness, its computational cost is prohibitive (over 2.5$\times$ slower than the baseline). The empirical 8-bin baseline provides robust multi-scale representations without introducing extreme computational overhead.

\begin{table*}[h]
    \centering
    \caption{Ablation on Shapley Sampling Budget. The baseline configuration is bolded.}
    \label{tab:ablation_budget}
    \begin{tabular}{@{}lccc@{}}
        \toprule
        \textbf{Budget ($N$)} & \textbf{Insertion AUC ($\uparrow$)} & \textbf{Deletion AUC ($\downarrow$)} & \textbf{Runtime (s)$^*$} \\ \midrule
        32            & 0.871          & 0.264          & 1.34 \\
        64            & 0.873          & 0.248          & 1.46 \\
        128           & 0.875          & 0.245          & 1.71 \\
        256           & 0.875          & 0.244          & 2.04 \\
        512           & 0.875          & 0.244          & 2.50 \\
        \textbf{1024 (Baseline)} & \textbf{0.875} & \textbf{0.244} & \textbf{3.09} \\
        2048          & 0.875          & 0.244          & 3.76 \\
        4096          & 0.875          & 0.244          & 4.64 \\ \bottomrule
        \multicolumn{4}{l}{\footnotesize $^*$Runtime includes full AUC curve evaluation overhead.}
    \end{tabular}
\end{table*}

\begin{table*}[h]
    \centering
    \caption{Ablation on Morphological Overlap Threshold. The baseline configuration is bolded.}
    \label{tab:ablation_overlap}
    \begin{tabular}{@{}lccc@{}}
        \toprule
        \textbf{Overlap Threshold} & \textbf{Insertion AUC ($\uparrow$)} & \textbf{Deletion AUC ($\downarrow$)} & \textbf{Runtime (s)$^*$} \\ \midrule
        0.10          & 0.854          & 0.249          & 1.68 \\
        0.20          & 0.857          & 0.255          & 2.06 \\
        0.30          & 0.863          & 0.254          & 2.33 \\
        0.40          & 0.871          & 0.250          & 2.71 \\
        \textbf{0.50 (Baseline)} & \textbf{0.875} & \textbf{0.244} & \textbf{3.09} \\
        0.60          & 0.882          & 0.234          & 4.08 \\
        0.70          & 0.889          & 0.224          & 5.01 \\
        0.80          & 0.895          & 0.218          & 6.30 \\
        0.90          & 0.902          & 0.213          & 7.59 \\ \bottomrule
        \multicolumn{4}{l}{\footnotesize $^*$Runtime includes full AUC curve evaluation overhead.}
    \end{tabular}
\end{table*}

\begin{table*}[h]
    \centering
    \caption{Ablation on Scale Granularity and Distribution. The baseline configuration is bolded.}
    \label{tab:ablation_scale}
    \begin{tabular}{@{}lccc@{}}
        \toprule
        \textbf{Scale Definition} & \textbf{Insertion AUC ($\uparrow$)} & \textbf{Deletion AUC ($\downarrow$)} & \textbf{Runtime (s)$^*$} \\ \midrule
        4 Bins (Linear)      & 0.737          & 0.455          & 1.02 \\
        10 Bins (Linear)     & 0.828          & 0.308          & 1.20 \\
        15 Bins (Linear)     & 0.853          & 0.275          & 1.58 \\
        \textbf{8 Bins (Baseline)} & \textbf{0.875} & \textbf{0.244} & \textbf{3.09} \\
        8 Bins (Log-spaced)  & 0.898          & 0.194          & 7.67 \\ \bottomrule
        \multicolumn{4}{l}{\footnotesize $^*$Runtime includes full AUC curve evaluation overhead.}
    \end{tabular}
\end{table*}

\paragraph{Robustness to Alternate Background Baselines.}
In our primary experiments, we utilize a fixed mid-gray background image. To ensure that our performance gains are not artifacts of this specific imputation method, we evaluated the robustness of \method across diverse background imputation strategies, including Gaussian blur, random uniform noise, pure white, pure black, and per-image global mean color. As shown in Table~\ref{tab:background_ablation}, \method remains highly robust across masking strategies. While using the image's global mean slightly optimizes the metrics (Insertion AUC: 0.882), the \textit{gray} baseline (0.875) provides excellent stability without requiring dynamic per-image tensor recalculations. Notably, the pure \textit{black} baseline degrades performance significantly (Insertion AUC drops to 0.786), primarily driven by severe out-of-distribution artifacts induced in the satellite imagery (EuroSAT) domain.

\begin{table*}[h]
    \centering
    \caption{Robustness of \method across different background imputation strategies, averaged across ImageNet and EuroSAT benchmarks.}
    \label{tab:background_ablation}
    \small
    \begin{tabular}{lcc}
        \toprule
        \textbf{Background Imputation Strategy} & \textbf{Insertion AUC ($\uparrow$)} & \textbf{Deletion AUC ($\downarrow$)} \\
        \midrule
        Global Image Mean & 0.882 & 0.252 \\
        \textbf{Mid-Gray (Baseline)} & \textbf{0.875} & \textbf{0.244} \\
        Uniform Noise & 0.852 & 0.197 \\
        Pure White & 0.839 & 0.186 \\
        Gaussian Blur & 0.822 & 0.192 \\
        Pure Black & 0.786 & 0.337 \\
        \bottomrule
    \end{tabular}
\end{table*}

\begin{table*}[h]
    \centering
    \caption{Comparative performance of geometry tagging approaches on isolated synthetic shapes versus components extracted end-to-end by the \method pipeline from complex synthetic composites.}
    \label{tab:geometry_tagger_benchmark}
    \small
    \begin{tabular}{l c cc}
        \toprule
        & \textbf{Isolated Synthetic} & \multicolumn{2}{c}{\textbf{Pipeline Extracted Components}} \\
        \cmidrule(lr){2-2} \cmidrule(lr){3-4}
        \textbf{Classifier Framework} & \textbf{Accuracy (\%)} & \textbf{Global Accuracy (\%)} & \textbf{Macro F1-Score} \\
        \midrule
        Rule-based geometric naming & 89.30\% & 59.83\% & 0.554 \\
        \textbf{Learned MLP geometric naming} & \textbf{99.60\%} & \textbf{77.86\%} & \textbf{0.777} \\
        \bottomrule
    \end{tabular}
\end{table*}

\paragraph{Isolated vs. Pipeline Resilience.}
As shown in Table~\ref{tab:geometry_tagger_benchmark}, under pristine isolated conditions, the feature-learned MLP achieves near-perfect classification accuracy (99.60\%). However, when deployed end-to-end on multi-shape composite images, the extracted morphological components inherently suffer from extraction noise, boundary discretization artifacts, and occlusion from overlapping shapes. Consequently, hand-crafted geometric rules drop sharply to 59.83\% global accuracy. In contrast, our feature-learned MLP maintains high resilience to this algorithmic extraction noise, achieving 77.86\% global accuracy across all extracted pipeline components, proving its necessity for downstream applications.

\paragraph{Real-World Geometry Recovery on GTSRB.}
To quantitatively evaluate whether \method's morphological decomposition successfully isolates true geometric primitives in noisy, natural imagery (overcoming background clutter and camera perspective skew), we conducted an empirical validation on the German Traffic Sign Recognition Benchmark (GTSRB)~\cite{Stallkamp2012}. Traffic signs possess legally standardized, ground-truth geometry (e.g., speed limits as \textit{Circles}, warning signs as \textit{Triangles}, and priority signs as \textit{Rectangles}). 

We evaluated $N=500$ test images spanning these classes. As detailed in Table~\ref{tab:gtsrb_geometry_eval}, the neural MLP tagger achieves an overall primitive recovery rate of \textbf{74.44\%}, significantly outperforming the rule-based heuristic (57.00\%). The accuracy is particularly robust for sharp, bounded geometry such as \textit{Triangles} (86.6\%) and \textit{Rectangles} (85.3\%). This confirms that \method's hierarchical decomposition successfully isolates macro-morphological boundaries, and that our MLP effectively maps warped contours to human-interpretable geometric tags.

\begin{table*}[h]
    \centering
    \caption{Real-world primitive recovery rate on GTSRB traffic sign contours across 500 test instances. A recovery is counted if the \method pipeline successfully extracts and correctly names the ground-truth geometric shape of the traffic sign.}
    \label{tab:gtsrb_geometry_eval}
    \small
    \begin{tabular}{l c cc}
        \toprule
        \textbf{Shape Class} & \textbf{Sample Size ($N$)} & \textbf{Rule-Based (\texttt{geom\_tagger})} & \textbf{Neural MLP (\texttt{mlp\_tagger})} \\
        \midrule
        Circle & 317 & 49.5\% & 67.8\% \\
        Triangle & 142 & 67.6\% & \textbf{86.6\%} \\
        Rectangle & 34 & 82.4\% & \textbf{85.3\%} \\
        \midrule
        \textbf{Overall Recovery Rate} & \textbf{493} & \textbf{57.00\%} & \textbf{74.44\%} \\
        \bottomrule
    \end{tabular}
\end{table*}

\section{Detailed User Study Setup and Results}
\label{Appendix:Detailed_User_Study}

\subsection{Study Setup}

\paragraph{Stimuli.}
We compiled a dataset of 150 images from the ImageNet Large Scale Visual Recognition Challenge 2012 (ILSVRC2012)~\cite{deng2009imagenet,ILSVRC15} validation dataset. Since most of the ImageNet classes are highly specific or technical, we used a Large Language Model (Google Gemini) to generate a broad, easily recognizable category for each label. From this augmented set, we randomly selected 150 unique classes and sampled one random image per class. To ensure participants could accurately identify the objects without requiring domain expertise, the stimuli were presented with the original label followed by the simplified category in parentheses (e.g. "Schipperke (Dog)", "Titi (Monkey)".

\paragraph{Participants.}
A total of 43 participants took the survey. There were no specific inclusion criteria. The recruited participants are from 7 countries across 3 continents.

\begin{figure*}[t]
	\centering
	\includegraphics[width=\linewidth]{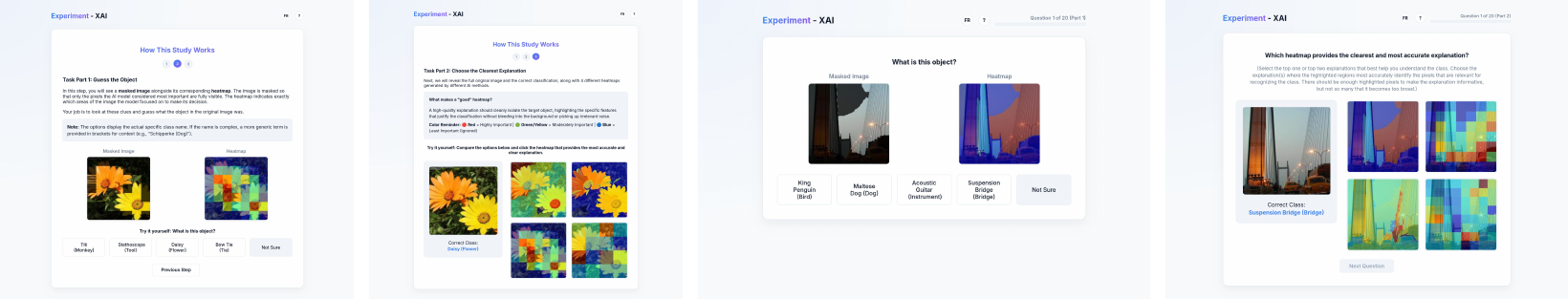}
	\caption{User Study Interface. From left to right: the first two images show the tutorial given to the users to explain the task. The last two images provide an example of the main interface.}
	\label{fig:survey_screenshots}
\end{figure*}

\paragraph{Study Procedure.}
Prior to the main evaluation, we provided a detailed tutorial to familiarize participants with the interface and the task (see Figure~\ref{fig:survey_screenshots}). Each participant inspected 20 images. For each image, we prepared heatmaps using four different methods: \method, PartitionSHAP (Axis Aligned Partitioning), Grad-CAM, and ShapBPT. 

The evaluation for each image consisted of two tasks. For the first task, participants were evaluated on a single heatmap and corresponding masked image. To ensure balanced representation per user, the assigned method was strictly controlled so that every participant saw exactly 5 heatmaps from each of the 4 methods across their 20 trials, with the presentation sequence fully randomized. For each image, users completed the following steps:
\begin{itemize}
    \item \textbf{Classification.} Participants were presented with a single heatmap and its corresponding masked image side-by-side and asked, "What is this object?". Their task was to identify the correct class of the image from five multiple-choice options (four specific classes and a fifth "Not sure" option). To ensure a fair comparison, the size of the masked region was strictly controlled across all methods. For \method, the mask was generated by selecting its most important morphological blob. For the pixel-attribution baselines (PartitionSHAP, Grad-CAM, and ShapBPT), we selected their top salient pixels until the total area exactly matched the area of the \method blob. This constraint guarantees that no baseline benefited from revealing a larger portion of the image.
    \item \textbf{Subjective Preference.} Immediately following the classification task, participants were shown the unmasked original image alongside its correct class label. All four generated heatmaps were displayed simultaneously. Participants were then asked, "Which heatmap provides the clearest and most accurate explanation?" To guide their evaluation, they were explicitly instructed to: (1) select the top one or two explanations that best helped them understand the class, and (2) choose explanations where the highlighted regions accurately identified relevant pixels without being so broad that they became uninformative.
\end{itemize}

\subsection{Participant Demographics}

\begin{figure*}[t]
	\centering
	\includegraphics[width=\linewidth]{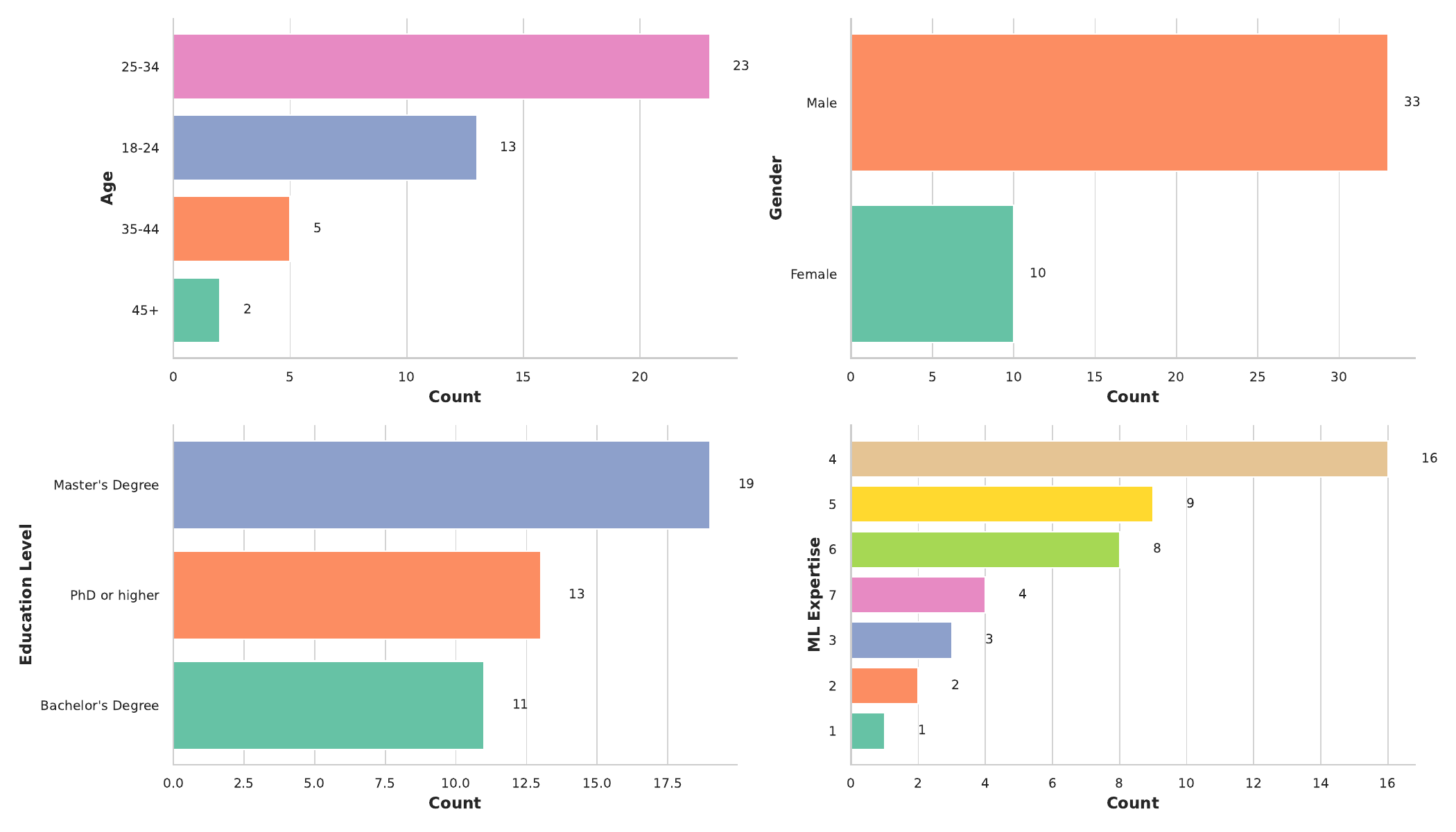}
	\caption{Participant Demographics including age, gender, education level and ML expertise.}
	\label{fig:demographics}
\end{figure*}

We collected responses from 42 participants from 7 unique countries. Figure~\ref{fig:demographics} provides a comprehensive overview of the participant demographics, highlighting the diversity in age, gender, education and AI expertise.

\subsection{Detailed Metrics, Statistical Analysis and Subgroup Analysis}

\begin{figure*}[t]
	\centering
	\includegraphics[width=\linewidth]{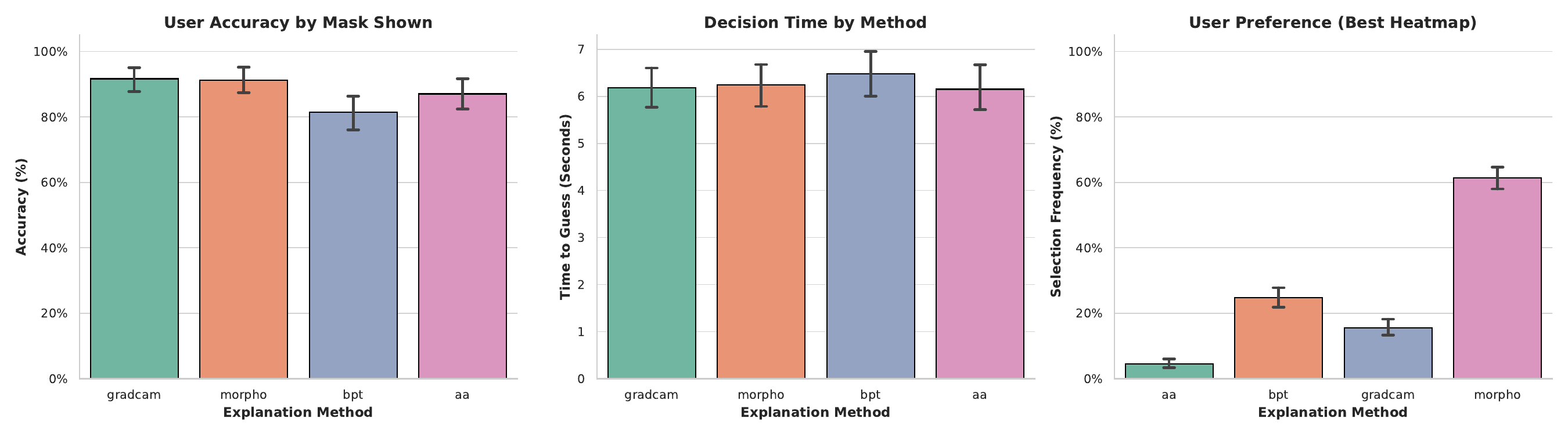}
	\caption{Detailed comparison of confidence and trust scores across methods.}
	\label{fig:survey_breakdown}
\end{figure*}

In addition to the overall method comparison presented in the main text, Figure~\ref{fig:survey_breakdown} shows detailed breakdowns of accuracy, decision time and preference.

\begin{figure*}[t]
	\centering
	\includegraphics[width=\linewidth]{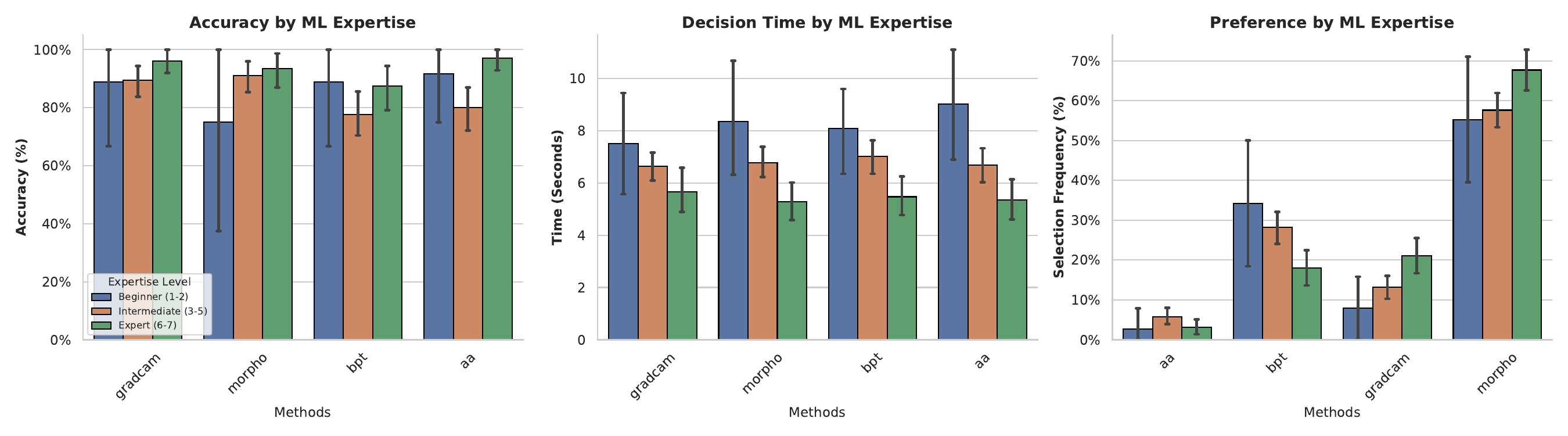}
	\caption{Performance Metrics categorized by participants' ML expertise.}
	\label{fig:demographic_ml_expertise}
\end{figure*}

\begin{figure*}[t]
	\centering
	\includegraphics[width=\linewidth]{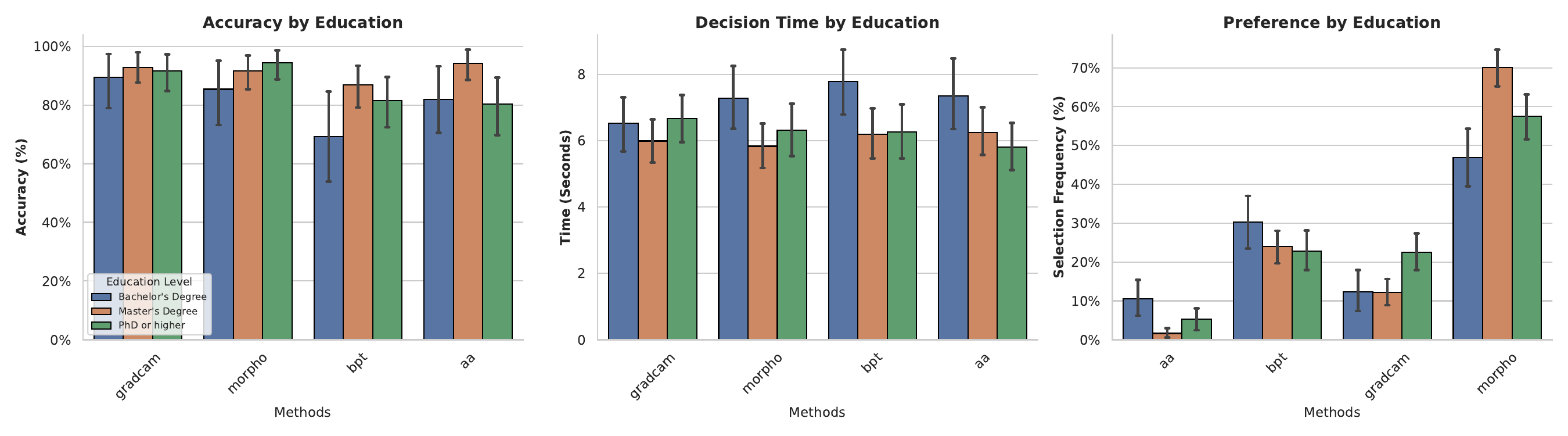}
	\caption{Performance Metrics categorized by participants' education levels.}
	\label{fig:demographic_education}
\end{figure*}

Furthermore, we analyzed the results across different demographic subgroups, as shown in Figures~\ref{fig:demographic_education} and~\ref{fig:demographic_ml_expertise}, indicating that \method's improvements hold consistently regardless of the participants ML expertise or educational background.

\paragraph{Statistical Analysis.} To validate our findings, we conducted a one-way Analysis of Variance (ANOVA) across the evaluated explanation methods. The analysis reports a statistically significant main effect of the explanation method on user accuracy ($F=4.4, p<0.01$). However, the effect of the method on user decision time was not found to be statistically significant ($F=0.4, p<0.75$). These results indicate that \method significantly improves the users' ability to correctly identify the target class without requiring additional cognitive processing time compared to baselines. Additionally, a Chi-Square goodness-of-fit test on the multi-select preference data confirmed that user selections for the best heatmap were not uniformly distributed ($\chi^2=552.4, p<0.001$), demonstrating a highly significant qualitative preference for \method over all baseline methods.

\end{document}